\documentclass[lettersize,journal]{IEEEtran}
\usepackage{amsmath,amsfonts}

\usepackage{algorithmic}
\usepackage{algorithm}
\usepackage{array}
\usepackage[caption=false,font=normalsize,labelfont=sf,textfont=sf]{subfig}
\usepackage{textcomp}
\usepackage{stfloats}
\usepackage{url}
\usepackage{verbatim}
\usepackage{graphicx}
\usepackage{cite}
\usepackage{graphicx}
\usepackage{amsmath}
\usepackage{amssymb}
\usepackage{booktabs}

\usepackage{multirow}
\usepackage{algorithm}
\usepackage{algorithmic}

\usepackage{setspace}

\usepackage{colortbl}
\usepackage{xcolor}
\usepackage{array} 
\usepackage{float}
\usepackage{arydshln}

\begin{document}

\title{Fooling the Image Dehazing Models by First Order Gradient}

\author{Jie Gui, Xiaofeng Cong, Chengwei Peng, Yuan Yan Tang, James Tin-Yau Kwok

        % <-this % stops a space
\thanks{J. Gui is with the School of Cyber Science and Engineering, Southeast University and with Purple Mountain Laboratories, Nanjing 210000, China (e-mail: guijie@seu.edu.cn).

X. Cong is with the School of Cyber Science and Engineering, Southeast University,  Nanjing 210000, China (e-mail: cxf\_svip@163.com).

C. Peng is with the Tecent Company, Shenzhen 518000, China (e-mail: cwpeng@tencent.com).

Y. Tang is with the Department of Computer and Information Science, University of Macau, Macau 999078, China (e-mail: yuanyant@gmail.com).

J. Kwok is with the Department of Computer Science and Engineering, The Hong Kong University of Science and Technology, Hong Kong 999077, China. (e-mail: jamesk@cse.ust.hk).
}}

\markboth{Journal of \LaTeX\ Class Files,~Vol.~14, No.~8, August~2021}%
{Shell \MakeLowercase{\textit{et al.}}: A Sample Article Using IEEEtran.cls for IEEE Journals}


% \makeatletter
% \def\ps@IEEEtitlepagestyle{
%   \def\@oddfoot{\mycopyrightnotice}
%   \def\@evenfoot{}
% }
% \def\mycopyrightnotice{
%   {\footnotesize
%   \begin{minipage}{\textwidth}
%   \centering
%   Copyright~\copyright~20xx IEEE. Personal use of this material is permitted. \\ 
%    However, permission to use this material for any other purposes must be obtained from the IEEE by sending a request to pubs-permissions@ieee.org.
%   \end{minipage}
%   }
% }

\maketitle

\begin{abstract}
The research on the single image dehazing task has been widely explored. However, as far as we know, no comprehensive study has been conducted on the robustness of the well-trained dehazing models. Therefore, there is no evidence that the dehazing networks can resist malicious attacks. In this paper, we focus on designing a group of attack methods based on first order gradient to verify the robustness of the existing dehazing algorithms. By analyzing the general purpose of image dehazing task, four attack methods are proposed, which are predicted dehazed image attack, hazy layer mask attack, haze-free image attack and haze-preserved attack. The corresponding experiments are conducted on six datasets with different scales. Further, the defense strategy based on adversarial training is adopted for reducing the negative effects caused by malicious attacks. In summary, this paper defines a new challenging problem for the image dehazing area, which can be called as adversarial attack on dehazing networks (AADN). Code and Supplementary Material are available at {https://github.com/Xiaofeng-life/AADN\_Dehazing}.
\end{abstract}


\section{Introduction}
\IEEEPARstart{T}{he} goal of image dehazing is to restore the clear scene from the hazy image~\cite{liu2022towards,yang2022self,song2022wsamf,wang2021tms,zhang2021hierarchical}, which is an important topic on low-level computer vision tasks. The image dehazing task~\cite{gui2022comprehensive} can be used as an upstream task of autonomous driving, traffic monitoring, and robot exploration, etc. Various dehazing networks have been proposed and verified on benchmark datasets~\cite{li2017aod,deng2019deep,dong2020multi,qin2020ffa,liu2019griddehazenet,zheng2021ultra}. An important question is, do these dehazing models behave securely during the inference stage of practical applications? Unfortunately, the security of dehazing algorithms~\cite{song2022wsamf,wang2021tms,zhang2021hierarchical} has not received due attention. In this paper, we find that well-trained dehazing networks with high performance can be fooled by the attack methods based on first order gradient~\cite{madry2017towards}. 

The existing dehazing networks~\cite{gui2022comprehensive} use different loss functions, network blocks and training strategies, but they may have two properties in common. First, the first-order gradient descent algorithm~\cite{gui2022comprehensive} is used for the training process. Second, one hazy image is taken as input and one dehazed image is provided as output without estimating the transmission map and atmospheric light in an explicit way during the inference stage. We call the dehazing methods that satisfy these two properties as the SISO (single input single output) dehazing networks. Within the scope of our study, we argue that SISO dehazing networks can be attacked by unified malicious attack algorithms.

%-------------------------------------------------------------------------
%------------------------------------------------------------------------
\begin{figure}
  \small
  \centering
  \includegraphics[width=8cm,height=4cm]{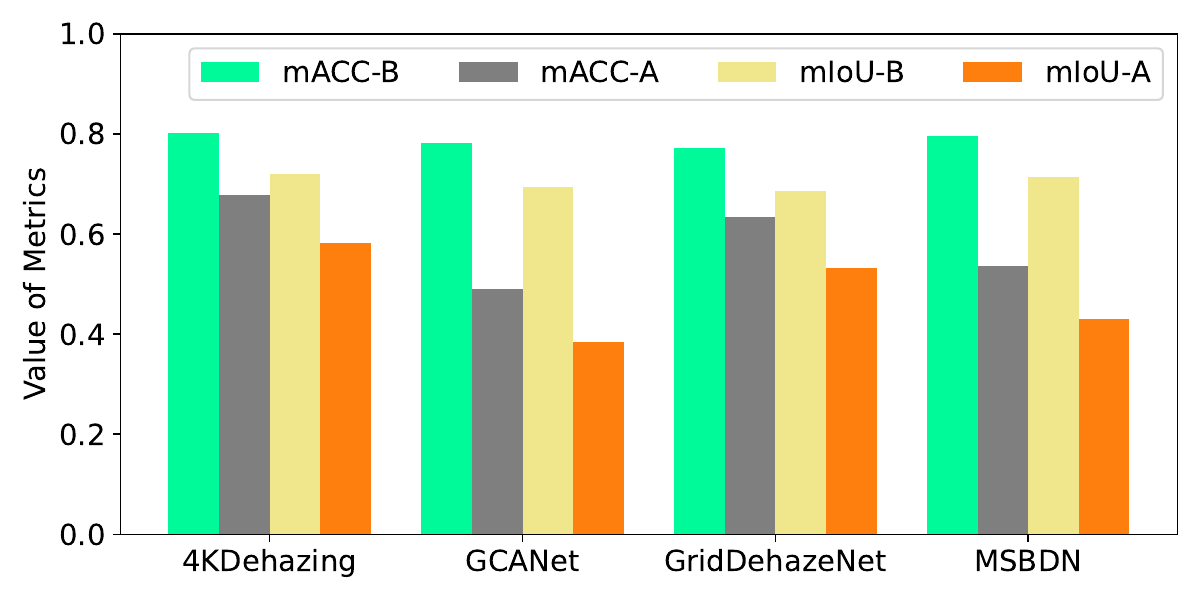}
  \caption{Attack results ($L_{P}^{MSE}$) on semantic segmentation dataset Foggy-City, where mACC and mIoU are two metrics for semantic segmentation. ``-B'' means before attack, and ``-A'' is after attack.}
  \label{fig:attack_quantitative_seg_cityscapes_before_defense}
\end{figure}

%-------------------------------------------------------------------------
%------------------------------------------------------------------------

The results in Figure \ref{fig:pred_mse_cityscapes_before_and_after_defense} show that after attacking the dehazing algorithm, the quality of the image is significantly reduced. Not only that, we found that the attack on the dehazing network will also cause the performance of the image segmentation algorithm, which may used as a downstream task of the dehazing task, to decline. Specifically, We choose the SegFormer~\cite{xie2021segformer} as the baseline segmentation network. The annotations for segmentation are from~\cite{cordts2016cityscapes}. During the training stage, the segmentation network is trained on the original clear images in the dataset. During the inference stage, the inputs of the segmentation network are the outputs of the dehazing network. Figure \ref{fig:attack_quantitative_seg_cityscapes_before_defense} provides the quantitative evaluation of the segmentation task on the whole test dataset. It is an obvious conclusion that the attack on dehazing networks will lead to worse segmentation results.

Specifically, we propose two questions for the adversarial attack on dehazing networks (AADN) research as follows.
\begin{itemize}
  \item Can the state-of-the-art dehazing models be attacked? What are the effective ways to attack them?
  \item If the dehazing models are vulnerable, how can we protect them in a proper way?
\end{itemize}

By obtaining the first-order gradient information, we successfully attack the dehazing models in a white-box manner. From the attacker's perspective, unlike the class-based image classification task, the outputs of dehazing network are in pixel-wise form. The main difficulty is that the attacker may not have the ``ground-truth label'' for attacking. Therefore, we propose an attack method based on the predicted dehazed image that can avoid using the ground-truth label. Meanwhile, the difference in attack performance obtained using the ground-truth label versus the predicted dehazed image is studied. Then, considering that haze is not always uniformly distributed across the image, we design an attack method based on the haze-layer, which can partially attack haze images. Further, we explored how to preserve the haze in the hazy image, which means that the input and output of the dehazing network are almost identical. After finding that the dehazing networks can be attacked by the first order attack algorithm, a natural question is whether we can design a corresponding defense method to improve the performance after being attacked. Inspired by the research on robust image classification~\cite{madry2017towards}, an adversarial training strategy is adopted for the defense process. The main contributions of this paper are as follows.

\begin{itemize}
  \item We define a new valuable research area called AADN. The study about adversarial attack and defense on dehazing task is conducted on various dehazing models and benchmark datasets.
  \item A pixel-wise white-box~\cite{akhtar2021advances,kherchouche2021detect} attack form is proposed for image dehazing task as the basic attack form, which is based on the first order gradient information during the backpropagation process.
  \item By analyzing the characteristics of SISO dehazing task, four attack methods are designed and verified on six benchmark datasets. The four attack methods are predicted dehazed image attack, hazy layer mask attack, haze-free image attack and haze-preserved attack, respectively.
  \item To explore if the attack effects can be reduced, the adversarial training is used for protecting the dehazing models with the target guiding way. Further, we find that adversarial defense training can improve the robustness of dehazing networks, but it can not ``perfectly'' protect the dehazing networks.
\end{itemize}

Our research shows that the SISO dehazing models can be attacked by the first-order gradient-based attack algorithm. Furthermore, we get three important conclusions. First, adversarial perturbations acquired with gradient information have significantly stronger negative effects on the dehazing model than regular noises. Second, the attacker only needs to use the outputs of dehazing models as pseudo-labels instead of using the ground-truth haze-free images as the guidance information for the attack process. Third, adversarial defense training can reduce the negative impacts caused by the attack algorithm, but it cannot completely eliminate these impacts. Meanwhile, adversarial defense training will slightly reduce the dehazing performance of the dehazing model on normal examples.

%-------------------------------------------------------------------------
%-------------------------------------------------------------------------
\begin{figure*}
  \centering
  \includegraphics[width=17cm,height=4.5cm]{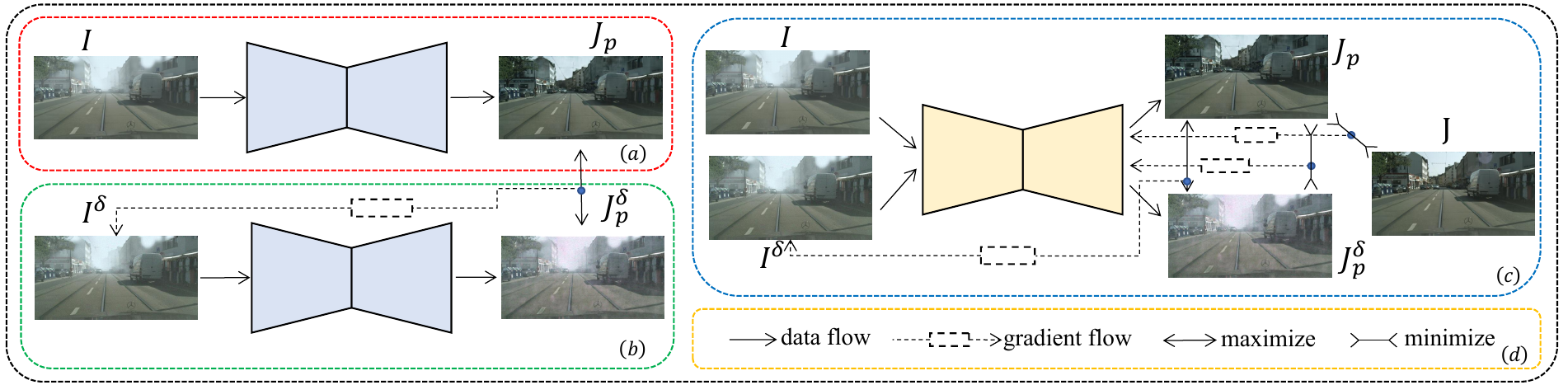}
  \caption{The pipelines of adversarial attack and defense on AADN. The contents of the four dotted boxes are (a) original dehazing process, (b) attack dehazing network, (c) adversarial defense training, and (d) illustration of arrows, respectively.}
  \label{fig:idea_of_paper}
\end{figure*}
%-------------------------------------------------------------------------
%-------------------------------------------------------------------------

\section{Related Work}
%-------------------------------------------------------------------------
%-------------------------------------------------------------------------
\subsection{Image Dehazing}
Image dehazing algorithms aim at removing the haze on the hazy images~\cite{liu2022towards,yang2022self,zhang2021single,li2021dehazeflow,li2022physically}. The existing dehazing methods can be divided into DL-based and non-DL-based~\cite{gui2022comprehensive}. Overall, without considering the computation cost, the DL-based dehazing methods can achieve better performance than non-DL-based methods. However, our research shows the DL-based SISO dehazing models can be attacked by the first order attack algorithms.

Various dehazing networks~\cite{gui2022comprehensive} have been proposed by the researchers. DM2FNet~\cite{deng2019deep} adopts four different feature functions and fuses them to the final output. The attention mechanism is applied in FFANet~\cite{qin2020ffa} and GridDehazeNet~\cite{liu2019griddehazenet}, which can obtain fine dehazing results. MSBDN~\cite{dong2020multi} proves that boosting strategy and back projection can be used in the dehazing module, which can preserve spatial information by a symmetrical path. The contrastive learning is used for dehazing task by AECR-Net~\cite{wu2021contrastive}, which constructs the positive and negative pairs during the training process. DIDH~\cite{shyam2021towards} proposes to restore the high frequency and low frequency information by discriminator networks. RDN~\cite{li2020deep} proves that retinex model can be embedded to the dehazing process to obtain the residual illumination map and the dehazed image. PSD~\cite{chen2021psd} takes a pre-trained and fine-tuning strategy to utilize the physical priors. DeHamer~\cite{guo2022image} uses the transformer architecture to implement the dehazing network. 4KDehazing~\cite{zheng2021ultra} puts affine bilateral grid learning as a branch to the overall network that can process the high resolution image with high speed. Within the scope of SISO dehazing system, we find these well-known dehazing networks can be fooled by first order attack.

%-------------------------------------------------------------------------
%-------------------------------------------------------------------------
\subsection{Adversarial Attack and Defense}
The adversarial attack and defense are important topics for artificial intelligence security~\cite{tramer2020adaptive,athalye2018obfuscated,uesato2018adversarial,kherchouche2020detection,chakraborty2018adversarial,wang2020you}. It is an interesting question to evaluate the robustness of neural networks under attack~\cite{carlini2017towards}. The research on adversarial attack argue that the well-trained networks may be fooled by hostile attack~\cite{yuan2019adversarial,deldjoo2021survey}. By adding a small perturbation to the example, the prediction label generated by the classification network may change from ``panda'' (correct) to ``gibbon'' (incorrect)~\cite{yuan2019adversarial}. 

Goodfellow proposes a fast gradient sign method (FGSM) ~\cite{goodfellow2014explaining} to generate the perturbation. FGSM is a one-step attack method which utilizes the gradient information calculated by the back propagation process. An iteration-based fast gradient sign method (i-FGSM) inspired by FGSM is explored by Kurakin et al.~\cite{kurakin2018adversarial}. Instead of generating the perturbation by one step backward like FGSM, i-FGSM update the perturbation by multi-step and clip it into a fixed range. Besides, projected gradient descent (PGD)~\cite{madry2017towards} is also a promising attack method to obtain the adversarial example that uses the multi-step update strategy. Inspired by PGD and ~\cite{yu2022towards}, the methods for attacking the dehazing networks are proposed in this paper.

Recent research has explored the topic on adversarial attack on image deraining~\cite{yu2022towards}, semantic segmentation~\cite{xie2017adversarial,nakka2020indirect}, super-resolution~\cite{castillo2021generalized,mustafa2019image,choi2019evaluating} and deblurring~\cite{choi2021deep}. \cite{castillo2021generalized} proposes a method that leverages the generalization capability of adversarial attacks to tackle real-world super-resolution models. Extensive empirical evaluations in \cite{mustafa2019image} show that image super-resolution can be an effective defense strategy against attack methods. Gao et al.~\cite{gao2021advhaze} propose an adversarial haze synthetic process based on the atmospheric scattering model to fool the classifiers. Kanwal et al.~\cite{kanwal2021person} study how to use the haze to reduce or improve the performance of the person re-identification task. Sun et al.~\cite{sun2022rethinking} propose a targeted adversarial attack to boost object detection performance after restoration. A new perspective is proposed by Sun's research, which combines image dehazing with downstream tasks. The goal of ~\cite{gao2021advhaze,kanwal2021person,sun2022rethinking} are different from this paper, since our purpose is to attack the dehazing networks. To the best of our knowledge, no comprehensive research on the dehazing adversarial attack has been published yet.

This paper focuses on the white-box attack. White-box attack, gray-box attack and black-box attack are popular research areas currently~\cite{akhtar2021advances,kherchouche2021detect,li2021simple}. White-box attack assumes the information of the networks can be obtained, like architecture and parameters. Gray-box attack and black-box attack assume the information is partially knowable and completely unknowable, respectively. In addition, ~\cite{pal2020game,madry2017towards} propose to treat the adversarial attack and defense as a game theoretic framework. In our experiments, we try to protect the dehazing network by adversarial training from the white-box attack.
%-------------------------------------------------------------------------
%-------------------------------------------------------------------------
\section{Methods}

The important concepts for AADN are as follows, which are consistent with the whole paper.
\begin{itemize}
  \item Attack: a process trying to reduce the performance of the well-trained dehazing networks by adding subtle perturbation to the hazy image.
  \item Attacker: a hostile attack algorithm, which can generate adversarial perturbation.
  \item Perturbation: a subtle signal generated by the attacker, and it is added to the hazy image.
  \item Adversarial example: a hazy image with perturbations.
  \item Defense: a process that tries to reduce the negative effects caused by the attacker.
\end{itemize}

The hazy image is denoted as $I$ and the corresponding ground-truth haze-free image is marked as $J$. $J_{p}$ stands for the predicted dehazed image obtained by dehazing network.

There are three important parts in this section. The basic attack form is introduced in Subsection \ref{subsec:basic_attack_form}, which defines the general way to attack the SISO dehazing system. Subsection \ref{subsec:how_to_attack} describes four attack methods to disturb the dehazing networks. Subsection \ref{subsec:how_to_defense} shows the adversarial training defense strategy to obtain robust dehazing networks. 
%-------------------------------------------------------------------------
%-------------------------------------------------------------------------
\subsection{Basic Attack Form}
\label{subsec:basic_attack_form}
First-order based gradient optimization methods \cite{gui2022comprehensive} are widely used for the SISO dehazing models. When attacking the dehazing model in a white-box manner, the gradient information is needed to calculate adversarial perturbations. Therefore, the first-order gradient information is adopted by our attack algorithm. The process of attacking the dehazing network is to add a perturbation $\delta$ to a hazy image $I$ and to obtain the adversarial example $I^{\delta}$, and this behavior makes the output ($J_{p}^{\delta}$) of dehazing network unpleasant, as follows,
\begin{equation}
  J_{p} = \Gamma_{\theta}(I),
\end{equation}
\begin{equation}
  \label{eq:generate_J_p_delta}
  J_{p}^{\delta} = \Gamma_{\theta}(I + \delta) = \Gamma_{\theta}(I^{\delta}),
\end{equation}
where $\Gamma_{\theta}$ denotes the pre-trained dehazing network with parameters $\theta$. The general optimization goal of image dehazing algorithms is to minimize one or more distance metrics, such as L1, L2, Perceptual~\cite{johnson2016perceptual} and structural similarity (SSIM)~\cite{wang2004image}. Therefore, the attacker tries to maximize these distance metrics, which will make the well-converged network have a higher loss value. The loss function $L_{att}$ that the attacker adopted is to maximize
\begin{equation}
  L_{att} = \Re(J_{p}^{\delta}, X),
\end{equation}
where the $\Re(\cdot, \cdot)$ represents various loss functions for image restoration process and $X$ is the target that the attacker needs to use. Different $X$ can be adopted by the attacker so that the dehazed image after the attack tends to have different forms of quality degradation. During the attack process, the parameters $\theta$ of the network $\Gamma_{\theta}$ will not be updated. The attacker needs to learn the perturbation $\delta$, which is updated by the gradient values
\begin{equation}
  \label{eq:delta_update_1}
  \delta^{t + 1} = \delta^{t} + \alpha sgn(\nabla_{\delta}L_{att}),
\end{equation}
where $sgn$ is the sign function, $\alpha$ is the adjusting factor for updating process, $\nabla$ denotes gradient operation, and the $t$ denotes $t$-th iteration step. For each iteration, the input $I^{\delta}$ of $\Gamma_{\theta}$ is not static, which is changed by the perturbation. To ensure that the perturbation $\delta$ added to hazy image $I$ can not be easily observed, the value range of $\delta$ should be controlled. As proposed in~\cite{madry2017towards}, $\delta$ can be constrained by $\ell_{\infty}$. Further, $\delta$ will be clipped as
\begin{equation}
  \label{eq:delta_update_2}
  \delta^{t + 1} = \kappa_{(-\epsilon, \epsilon)}(\delta^{t + 1}),
\end{equation}
where the $\kappa$ stands for clip operation and $\epsilon$ represents the range for clipping. $\epsilon = 0$ denotes no attack. The pixel values of all images are normalized to $[0, 1]$ during the training and inference stage. Therefore, the value range of attacked hazy image $I^{\delta}$ should also be limited to $[0, 1]$, so $\delta$ is further clipped by
\begin{equation}
  \label{eq:delta_update_3}
  \delta^{t + 1} = \kappa_{({0 - I, 1 - I})}(\delta^{t + 1}).
\end{equation}

The above attack form can be used in both targeted and non-targeted ways. The aim of the targeted attack is forcing $J_{p}^{\delta}$ close to $X$. The goal of the non-targeted attack is trying to make $J_{p}^{\delta}$ far from $J$ visually.

%-------------------------------------------------------------------------
%-------------------------------------------------------------------------
\subsection{How to Attack}
\label{subsec:how_to_attack}
Subsection \ref{subsec:basic_attack_form} gives the basic form about how to attack the dehazing networks. We propose that three important concepts that may be considered in the security of the dehazing task, namely the predicted dehazed image, the hazy image and the clear image. An attacker may use predicted dehazed images, hazy images and clear images to attack the dehazing models, respectively. When an attacker uses a predicted dehazed image or a clear image to attack the dehazing model, the output after the attack may deviate from the original output. When an attacker uses a hazy image to attack the dehazing model, the output after the attack may retain the haze in the image. Meanwhile, an attacker may perform partial attacks on the image. Considering the above factors, we designed four attack algorithms, which are predicted dehazed image attack $L_{P}$, hazy layer mask attack $L_{M}$, haze-free image attack $L_{G}$ and haze-preserved attack $L_{I}$, respectively.
%-------------------------------------------------------------------------
%-------------------------------------------------------------------------
\subsubsection{Predicted Dehazed Image Attack} 
Unlike image classification or semantic segmentation tasks, the ground-truth labels are not easy to be collected in the dehazing task. The hazy and haze-free pairs may not be annotated by humans. Therefore, we design an attack method based on predicted dehazed image $J_{p}$ that is obtained by the dehazing network $\Gamma_{\theta}$. The loss of the attack can be expressed by
\begin{equation}
  \label{eq:att_pred_J}
  L_{P} = \Re(\Gamma_{\theta}(I^{\delta}), J_{p}),
\end{equation}
where $P$ in $L_{P}$ denotes prediction. The optimization goal is to maximize $L_{P}$ in a non-targeted way. Since such an attack method is non-targeted, a natural question is whether $L_{P}$ will cause the prediction result $\Gamma_{\theta}(I^{\delta})$ after the attack to be closer to the ground-truth image and whether it will lead to a failure attack. The experiments show that $L_{P}$ is a reliable and powerful attack method. The original dehazing process is shown in Figure \ref{fig:idea_of_paper} (a), and the process of attacking dehazing networks is displayed in Figure \ref{fig:idea_of_paper} (b).
%-------------------------------------------------------------------------
%-------------------------------------------------------------------------
%-------------------------------------------------------------------------
%-------------------------------------------------------------------------
\begin{figure}
  \centering
  \footnotesize
  \includegraphics[width=8cm,height=4cm]{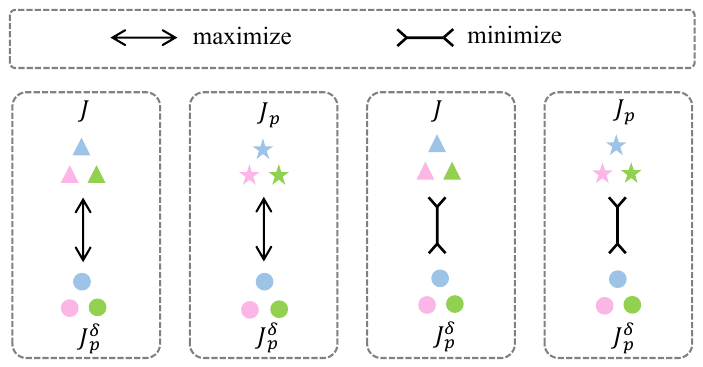}
  \leftline{\hspace{0.9cm} (a) $L_{G}$ \hspace{1.0cm} (b) $L_{P}$ \hspace{0.9cm} (c) $L_{def}^{G}$ \hspace{0.75cm} (d) $L_{def}^{P}$}
  \caption{The optimization direction of attack and defense.}
  \label{fig:idea_of_defense}
\end{figure}
%-------------------------------------------------------------------------
%-------------------------------------------------------------------------

\subsubsection{Haze Layer Mask Attack}
Equation (\ref{eq:att_pred_J}) provides an attack way based on the whole image. An interesting question is whether we can attack part of the hazy image and achieve impressive attack results. According to the atmospheric scattering model~\cite{mccartney1976optics}, the haze density of the pixels that are far from the camera is higher. This means that the haze is not always uniformly distributed across the image. Therefore, we propose the haze layer mask attack method which can partly attack the haze image, and the corresponding haze layer mask can be obtained by
\begin{equation}
  \mu = \frac{1}{H \times W} \sum_{m=1}^{H} \sum_{n=1}^{W} (I(m, n) - J_{p}(m, n)),
\end{equation}
\begin{equation}
  \label{eq:mask_attack}
  mask = \left\{
    \begin{aligned}
    1 \quad (I(m, n) - J_{p}(m, n)) > \mu, \\
    0 \quad (I(m, n) - J_{p}(m, n)) \leq \mu, \\
    \end{aligned}
    \right
    .
  \end{equation}
where $H$ and $W$ denotes the height and width of the image, respectively. The loss function of haze layer mask attack is to maximize
\begin{equation}
  L_{M} = \Re(\Gamma_{\theta}(I + \delta * mask), J_{p}),
\end{equation}
where $*$ denotes pixel-wise multiplication and $M$ in $L_{M}$ stands for the mask.

%-------------------------------------------------------------------------
%-------------------------------------------------------------------------
\subsubsection{Haze-free Image Attack}
Supposing the attacker already has the haze-free (ground-truth) image $J$, then the $J_{p}$ in attack loss $L_{P}$ can be replaced with $J$. The attack loss calculated by $J$ is 
\begin{equation}
L_{G} = \Re(\Gamma_{\theta}(I^{\delta}), J),
\end{equation}
where $G$ means ground-truth.
%-------------------------------------------------------------------------
%-------------------------------------------------------------------------
\subsubsection{Haze-preserved Attack}
Considering an extreme situation that the output of the dehazing model totally contains the haze in the original hazy image. In this situation, the optimization object of the attacker is to minimize the distance between the dehazed image and the original hazy image by $\Re(\Gamma_{\theta}(I^{\delta}), I)$. We multiply $\Re(\Gamma_{\theta}(I^{\delta}), I)$ by ``-1'' so that minimizing $\Re(\Gamma_{\theta}(I^{\delta}), I)$ is equivalent to maximizing $-1 * \Re(\Gamma_{\theta}(I^{\delta}), I)$, which is also equivalent to $- \Re(\Gamma_{\theta}(I^{\delta}), I)$. In this form, all attack loss functions proposed in this paper are optimized in a maximizing manner. The loss function for the haze-preserved attack is to maximize the following loss function
\begin{equation}
  L_{I} = -1 * \Re(\Gamma_{\theta}(I^{\delta}), I),
\end{equation}
where the subscript in $L_{I}$ means identity output.

%-------------------------------------------------------------------------
%-------------------------------------------------------------------------
\subsubsection{The Distance Metrics}
Here we choose two widely used distance metrics for $\Re(\cdot, \cdot)$, which are mean square error (MSE) and SSIM. The forms of MSE and SSIM are
\begin{equation}
  MSE(x, y) = ||x - y||_{2}^{2},
\end{equation} 
\begin{equation}
  SSIM(x, y) = \frac{(2\mu_{x}\mu_{y} + C1)(2\delta_{xy} + C2)}{(\mu_{x}^{2} + \mu_{y}^{2} + C1)(\delta_{x}^{2} + \delta_{y}^{2} + C2)},
\end{equation}
where $x$ and $y$ represent two images. C1 and C2 are constants. Details about SSIM can be found at~\cite{wang2004image}. The $\mu_{x} \& \mu_{y}$ and $\delta_{x} \& \delta_{y}$ denote mean and standard deviation, respectively. $\delta_{xy}$ is covariance. For convenience, the $L_{P}^{MSE}$ and $L_{P}^{SSIM}$ stand for the $L_{P}$ that using $MSE(x, y)$ and $1 - SSIM(x, y)$ as the loss function, respectively.

%-------------------------------------------------------------------------
%-------------------------------------------------------------------------
%-------------------------------------------------------------------------
%-------------------------------------------------------------------------
\begin{algorithm}
  \caption{The maximum number of iterations is set to $N$. The learning rate is denoted as $lr$. The iteration step of the attack process is $k$. $m$ is batch size. The boolean symbol $Tc$ is set to $True$ or $False$.}
  \label{alg:adv_training_defense}
  \begin{algorithmic}
  \STATE Initialize the dehazing network $\Gamma_{\theta}$ by well-trained parameters.
  \IF {$Tc$}
  \STATE Initialize the Teacher network $\Gamma_{\theta}^{T}$.
  \ENDIF
  \FOR{the maximum number of iterations $N$}
  \STATE Sample hazy images $I = \{i^{(1)}, i^{(2)}, \ldots, i^{(m)}\}$ and corresponding ground-truth images $J = \{j^{(1)}, j^{(2)}, \ldots, j^{(m)}\}$;
  \IF {$Tc$}
    \STATE Calculate the corresponding prediction $J_{p}^{T} = \{j_{p}^{T(1)}, j_{p}^{T(2)}, \ldots, j_{p}^{T(m)}\}$ by 
    \setstretch{0.5}
    \STATE
    \begin{equation}
      \label{eq:cal_pred_J}
      J_{p}^{T} = \Gamma_{\theta}^{T}(I).
    \end{equation}
  \ENDIF

  \STATE Initialize $\delta$ by uniform distribution $U(-\epsilon, \epsilon)$ and $\delta$ is clipped by Equation 6;
  \FOR{$k$ steps}
  \setstretch{1}

  \IF {$Tc$}
    \STATE Calculate the attack loss $L_{P}$;
  \ELSE
    \STATE Calculate the attack loss $L_{G}$;
  \ENDIF

  \STATE Update $\delta$ by Equations 4, 5 and 6.
  \ENDFOR
  \STATE Generate the attacked prediction $J_{p}^{\delta}$ by Equation 2;

  \IF {$Tc$}
  \STATE Calculate the adversarial training loss $L_{def}^{P}$;
  \ELSE
  \STATE Calculate the adversarial training loss $L_{def}^{G}$;
  \ENDIF

  \STATE Update parameters $\theta$ for $\Gamma_{\theta}$:
  \STATE
  \setstretch{0.5}
  \begin{equation}
  \theta = \theta - lr \times \nabla_{\theta}{L_{def}^{P}}.
  \end{equation}
  \ENDFOR
  \end{algorithmic}
  \end{algorithm}
%-------------------------------------------------------------------------
%-------------------------------------------------------------------------
%-------------------------------------------------------------------------
%-------------------------------------------------------------------------

\subsection{How to Defense}
\label{subsec:how_to_defense}

The attacker that adopts the white-box attack method can reduce the performance of the well-trained dehazing models. Therefore, we need to design the corresponding defense method to protect the vulnerable networks. There are two factors that we mainly consider.
\begin{itemize}
  \item The defense method can be adopted by different dehazing models, rather than designed for a specific one.
  \item The inference time should not be increased by the defense method.
\end{itemize} 

Based on the above two factors, we conduct the research on adversarial training defense~\cite{madry2017towards} to protect the dehazing networks. However, the method in ~\cite{madry2017towards} cannot be directly applied to the defense training of dehazing networks. The reason is that, as we analyzed when designing the attack algorithm, the target $X$ adopted by the attacker may not be a ground-truth image.

From the perspective of defense, the adversarial training process against the four attack methods can be divided into two categories. The first category is to use the original dehazed image $J_{p}$ to reduce the distance between the $J_{p}^{\delta}$ and $J_{p}$. Therefore, the adversarial training process may need to use the pre-trained dehazing model as a Teacher network to produce the original dehazed image $J_{p}$. This situation can be applied to $L_{P}$ and $L_{M}$. Figure \ref{fig:idea_of_defense} shows the optimization directions of attack and defense. From the attacker's perspective, when using attack loss $L_{P}$, the distance between $J_{p}^{\delta}$ and $J_{p}$ is maximized as shown in Figure \ref{fig:idea_of_defense}-(b). Therefore, theoretically, the defender should minimize the distance between $J_{p}^{\delta}$ and $J_{p}$ as shown in Figure \ref{fig:idea_of_defense}-(d), since the gradient direction of Figure \ref{fig:idea_of_defense}-(b) and Figure \ref{fig:idea_of_defense}-(d) are complementary. However, the process shown in Figure \ref{fig:idea_of_defense}-(d) requires the use of a fixed pre-trained model to calculate the $J_{p}$, which will increase the computational cost of the defense algorithm. Therefore, a possible way to eliminate the additional computational cost is to use the defense method shown in Figure \ref{fig:idea_of_defense}-(c) (which is designed for Figure \ref{fig:idea_of_defense}-(a)) to reduce the distance between $J_{p}^{\delta}$ and $J$. However, the gradient direction of Figure \ref{fig:idea_of_defense}-(b) and Figure \ref{fig:idea_of_defense}-(c) may not be complementary. Whether this way is feasible will be discussed and verified in the experimental section. The second category is to use clear image $J$ to reduce the distance between $J_{p}^{\delta}$ and $J$. This situation can be applied to $L_{J}$ and $L_{I}$.

Overall, two purposes are included in our research on defense training, as follows.

\begin{itemize}
  \item When using Teacher network or ground-truth images for defense training, can we effectively resist the corresponding attack algorithm?
  \item Can the use of the Teacher network be avoided to reduce computing costs?
\end{itemize}

Based on the above purposes, two adversarial defense training processes are designed as shown in Algorithm \ref{alg:adv_training_defense}. The attack methods are $L_{P}$ and $L_{G}$, respectively. In Algorithm \ref{alg:adv_training_defense}, the boolean type variable $Tc$ is used to represent whether the Teacher network is utilized. When the value of $Tc$ is $True$, the calculated attack and defense losses are $L_{P}$ and $L_{def}^{P}$, respectively. On the contrary, when the value of $Tc$ is $False$, the calculated attack and defense losses are $L_{G}$ and $L_{def}^{G}$, respectively. Other attack methods can use Algorithm \ref{alg:adv_training_defense} for their own adversarial training processes. The adversarial training defense for $L_{P}$ and $L_{G}$ is a min-max (mm) game, as follows,
\begin{equation}
  \label{eq:L_mm_P}
 L_{mm}^{P} = \min_{\theta} \Re(J_{p}^{\delta}, J_{p}^{T}) \max_{||\delta||_{\infty} \leq \epsilon} \Re(J_{p}^{\delta}, J_{p}^{T}),
\end{equation}
\begin{equation}
 L_{mm}^{G} = \min_{\theta} \Re(J_{p}^{\delta}, J) \max_{||\delta||_{\infty} \leq \epsilon} \Re(J_{p}^{\delta}, J).
\end{equation}

Since the parameters $\theta$ of dehazing network $\Gamma_{\theta}$ will be changed during the adversarial training process, the Teacher network $\Gamma_{\theta}^{T}$ with fixed parameters is adopted to obtain predicted dehazed image $J_{p}^{T}$ ($J_{p}^{T} = \Gamma_{\theta}^{T}(I)$) for $L_{def}^{P}$. To ensure the dehazing network can still handle the hazy images $I$ that have not been attacked, the basic image restoration loss function should be adopted for the final defense loss. The  overall losses of the corresponding defense methods for $L_{P}$ and $L_{G}$ are $L_{def}^{P}$ and $L_{def}^{G}$, which are defined as
\begin{equation}
  \label{eq:L_def_P}
  L_{def}^{P} = \Re(\Gamma_{\theta}(I), J) + \lambda L_{mm}^{P},
\end{equation}
\begin{equation}
  \label{eq:L_def_G}
  L_{def}^{G} = \Re(\Gamma_{\theta}(I), J) + \lambda L_{mm}^{G},
\end{equation}
where $\lambda$ is the balance factor for the original image restoration loss function and adversarial defense loss function. The adversarial training defense process does not need to start from scratch. The parameters can be initialized by the well-trained network, and fine-tuned on the adversarial examples. The overall pipeline about adversarial training defense $L_{def}^{P}$ is shown in Figure \ref{fig:idea_of_paper}-(c). 

%-------------------------------------------------------------------------
%------------------------------------------------------------------------
\begin{table*}
  \scriptsize
  \centering
  \caption{Quantitative attack results obtained by $L_{P}^{MSE}$, $L_{M}^{MSE}$ and $L_{I}^{MSE}$. The ``Method-$\epsilon$'' denotes $\epsilon=i, i \in \{2, 4, 6, 8\}$. The worst attack result and best attack result in each dehazing method are in \textcolor{blue}{blue} and \textcolor{red}{red}, respectively. The dehazing results without attack are in \textbf{bold}.}
  \label{tab:attack_results_differ_epsilon_by_LP_LM_LI}
  
  \begin{tabular}{c|c@{\hspace{0.27cm}}c|c@{\hspace{0.27cm}}c|c@{\hspace{0.27cm}}c|c@{\hspace{0.27cm}}c|c@{\hspace{0.27cm}}c|c@{\hspace{0.27cm}}c|c@{\hspace{0.27cm}}c}
    \hline
      \multirow{2}{*}{Methods} & \multicolumn{2}{c|}{ITS} & \multicolumn{2}{c|}{OTS} & \multicolumn{2}{c|}{4K} & \multicolumn{2}{c|}{Foggy-City} & \multicolumn{2}{c|}{I-HAZE} & \multicolumn{2}{c|}{O-HAZE} & \multicolumn{2}{c}{D-Hazy}\\
    \cline{2-15}
     & PSNR$\uparrow$ & SSIM$\uparrow$ & PSNR$\uparrow$ & SSIM$\uparrow$ & PSNR$\uparrow$ & SSIM$\uparrow$ & PSNR$\uparrow$ & SSIM$\uparrow$ & PSNR$\uparrow$ & SSIM$\uparrow$ & PSNR$\uparrow$ & SSIM$\uparrow$ & PSNR$\uparrow$ & SSIM$\uparrow$ \\

    \hline                     
      4KDehazing         & \textbf{26.780} & \textbf{0.927} & \textbf{28.724} & \textbf{0.956} & \textbf{21.634} & \textbf{0.891} & \textbf{31.461} & \textbf{0.974} & \textbf{16.826} & \textbf{0.684} & \textbf{20.916} & \textbf{0.769} & \textbf{23.304} & \textbf{0.875} \\
      \hdashline
      4KDehazing-P2      & 14.700 & 0.724 & 16.815 & 0.798 & 13.153 & 0.700 & 15.626 & 0.817 & 16.686 & 0.614 & 20.129 & 0.736 & 10.302 & 0.558 \\
      4KDehazing-P4      & 12.465 & 0.645 & 14.385 & 0.709 & 8.985  & 0.508 & 13.538 & 0.739 & 16.244 & 0.541 & 18.818 & 0.660 & 9.000  & 0.504 \\
      4KDehazing-P6      & 11.291 & 0.586 & 13.709 & 0.655 & 7.332  & 0.384 & 12.768 & 0.683 & 15.737 & 0.477 & 17.707 & 0.590 & 8.475  & 0.448 \\
      \hdashline
      
      4KDehazing-P8      & \textcolor{red}{10.545} & \textcolor{red}{0.533} & \textcolor{red}{13.181} & \textcolor{red}{0.605} & \textcolor{red}{6.617}  & \textcolor{red}{0.315} & \textcolor{red}{12.355} & \textcolor{red}{0.634} & \textcolor{red}{15.235} & \textcolor{red}{0.426} & \textcolor{red}{16.744} & \textcolor{red}{0.533} & \textcolor{red}{8.061}  & \textcolor{red}{0.426} \\
      4KDehazing-M8      & 12.213 & 0.631 & 15.697 & 0.745 & 9.177  & 0.586 & 13.845 & \textcolor{blue}{0.780} & 15.811 & 0.512 & \textcolor{blue}{17.427} & 0.600 & 9.597  & \textcolor{blue}{0.582} \\
      4KDehazing-I8      & \textcolor{blue}{12.725} & \textcolor{blue}{0.648} & \textcolor{blue}{16.470} & \textcolor{blue}{0.761} & \textcolor{blue}{12.001} & \textcolor{blue}{0.645} & \textcolor{blue}{14.900} & 0.709 & \textcolor{blue}{15.834} & \textcolor{blue}{0.587} & 17.274 & \textcolor{blue}{0.695} & \textcolor{blue}{9.985}  & 0.568 \\

      \hline
      \hline
   
      DM2FNet            & \textbf{29.922} & \textbf{0.954} & \textbf{29.323} & \textbf{0.957} & \textbf{26.651} & \textbf{0.943} & \textbf{32.582} & \textbf{0.971} & \textbf{17.368} & \textbf{0.711} & \textbf{23.519} & \textbf{0.754} & \textbf{28.252} & \textbf{0.937} \\
      \hdashline
      DM2FNet-P2         & 13.782 & 0.681 & 17.343 & 0.802 & 11.769 & 0.687 & 15.978 & 0.821 & 15.372 & 0.661 & 16.059 & 0.604 & 8.385  & 0.404 \\
      DM2FNet-P4         & 11.912 & 0.610 & 15.160 & 0.709 & 9.288  & 0.564 & 14.694 & 0.762 & 13.999 & 0.595 & 12.435 & 0.508 & 7.100  & 0.296 \\
      DM2FNet-P6         & 11.423 & 0.593 & 14.596 & 0.657 & 8.856  & 0.511 & 14.275 & 0.708 & 13.916 & 0.561 & 11.454 & 0.475 & 6.823  & 0.274 \\

      \hdashline
      DM2FNet-P8         & \textcolor{red}{11.214} & \textcolor{red}{0.570} & \textcolor{red}{14.464} & \textcolor{red}{0.617} & \textcolor{red}{8.624}  & \textcolor{red}{0.476} & \textcolor{red}{14.010} & \textcolor{red}{0.650} & \textcolor{red}{13.265} & \textcolor{red}{0.495} & \textcolor{red}{10.899} & \textcolor{red}{0.454} & \textcolor{red}{6.813}  & \textcolor{red}{0.268} \\
      DM2FNet-M8         & 11.983 & \textcolor{blue}{0.652} & 15.766 & 0.736 & 10.397 & \textcolor{blue}{0.644} & \textcolor{blue}{14.874} & \textcolor{blue}{0.792} & 14.338 & \textcolor{blue}{0.608} & 12.485 & 0.539 & 8.746  & 0.540 \\
      DM2FNet-I8         & \textcolor{blue}{12.207} & 0.644 & \textcolor{blue}{16.285} & \textcolor{blue}{0.748} & \textcolor{blue}{12.367} & 0.634 & 14.852 & 0.754 & \textcolor{blue}{14.681} & 0.558 & \textcolor{blue}{15.541} & \textcolor{blue}{0.640} & \textcolor{blue}{9.729}  & \textcolor{blue}{0.582} \\
    \hline
    \hline
  
      FFANet             & \textbf{29.414} & \textbf{0.962} & \textbf{29.839} & \textbf{0.959} & \textbf{24.641} & \textbf{0.936} & \textbf{35.053} & \textbf{0.981} & \textbf{16.507} & \textbf{0.749} & \textbf{22.587} & \textbf{0.853} & \textbf{27.441} & \textbf{0.919} \\
      \hdashline
      FFANet-P2          & 14.450 & 0.682 & 18.130 & 0.822 & 13.027 & 0.678 & 14.805 & 0.784 & 16.432 & 0.680 & 21.548 & 0.816 & 8.959  & 0.461 \\
      FFANet-P4          & 11.498 & 0.523 & 15.911 & 0.735 & 10.473 & 0.558 & 13.312 & 0.654 & 16.250 & 0.567 & 19.989 & 0.737 & 7.347  & 0.312 \\
      FFANet-P6          & 10.528 & 0.474 & 14.657 & 0.663 & 9.199  & 0.477 & 11.752 & 0.486 & 15.990 & 0.472 & 18.498 & 0.655 & 6.885  & 0.255 \\

      \hdashline
      FFANet-P8          & \textcolor{red}{10.091} & \textcolor{red}{0.448} & \textcolor{red}{13.978} & \textcolor{red}{0.611} & \textcolor{red}{8.742}  & \textcolor{red}{0.441} & \textcolor{red}{9.728}  & \textcolor{red}{0.326} & \textcolor{red}{15.664} & \textcolor{red}{0.400} & \textcolor{red}{17.487} & \textcolor{red}{0.579} & \textcolor{red}{6.556}  & \textcolor{red}{0.223} \\
      FFANet-M8          & 11.553 & 0.586 & 15.514 & 0.735 & 11.766 & \textcolor{blue}{0.646} & 12.403 & 0.597 & \textcolor{blue}{15.968} & 0.536 & 19.053 & 0.648 & 8.841  & 0.457 \\
      FFANet-I8          & \textcolor{blue}{12.386} & \textcolor{blue}{0.653} & \textcolor{blue}{16.608} & \textcolor{blue}{0.771} & \textcolor{blue}{12.805} & 0.613 & \textcolor{blue}{14.899} & \textcolor{blue}{0.718} & 15.690 & \textcolor{blue}{0.688} & \textcolor{blue}{19.412} & \textcolor{blue}{0.717} & \textcolor{blue}{10.338} & \textcolor{blue}{0.528} \\

    \hline
    \hline

      GCANet             & \textbf{25.880} & \textbf{0.895} & \textbf{27.808} & \textbf{0.943} & \textbf{23.266} & \textbf{0.880} & \textbf{31.541} & \textbf{0.968} & \textbf{16.535} & \textbf{0.669} & \textbf{20.599} & \textbf{0.717} & \textbf{24.209} & \textbf{0.881} \\
      \hdashline
      GCANet-P2          & 11.879 & 0.537 & 14.744 & 0.719 & 13.274 & 0.682 & 14.901 & 0.787 & 16.565 & 0.632 & 18.211 & 0.680 & 9.665  & 0.557 \\
      GCANet-P4          & 8.981  & 0.359 & 10.760 & 0.527 & 9.798  & 0.537 & 12.006 & 0.652 & 14.571 & 0.547 & 16.686 & 0.609 & 8.203  & 0.458 \\
      GCANet-P6          & 7.925  & 0.301 & 9.336  & 0.448 & 8.574  & 0.459 & 10.931 & 0.568 & 14.106 & 0.484 & 14.967 & 0.512 & 7.395  & 0.392 \\

      \hdashline
      GCANet-P8          & \textcolor{red}{7.530}  & \textcolor{red}{0.283} & \textcolor{red}{8.733}  & \textcolor{red}{0.407} & \textcolor{red}{7.990}  & \textcolor{red}{0.409} & \textcolor{red}{10.433} & \textcolor{red}{0.516} & \textcolor{red}{13.353} & \textcolor{red}{0.424} & \textcolor{red}{14.200} & \textcolor{red}{0.455} & \textcolor{red}{6.932}  & \textcolor{red}{0.354} \\
      GCANet-M8          & 9.485  & 0.453 & 10.803 & 0.598 & 10.050 & 0.554 & 12.971 & \textcolor{blue}{0.742} & 14.409 & 0.540 & 15.389 & 0.519 & 8.410  & 0.502 \\
      GCANet-I8          & \textcolor{blue}{13.433} & \textcolor{blue}{0.645} & \textcolor{blue}{17.739} & \textcolor{blue}{0.755} & \textcolor{blue}{12.010} & \textcolor{blue}{0.599} & \textcolor{blue}{15.656} & 0.726 & \textcolor{blue}{15.843} & \textcolor{blue}{0.585} & \textcolor{blue}{17.659} & \textcolor{blue}{0.600} & \textcolor{blue}{10.492} & \textcolor{blue}{0.563} \\

    \hline
    \hline 

      GridDehazeNet      & \textbf{23.243} & \textbf{0.934} & \textbf{27.806} & \textbf{0.961} & \textbf{22.614} & \textbf{0.935} & \textbf{30.508} & \textbf{0.961} & \textbf{15.872} & \textbf{0.676} & \textbf{19.543} & \textbf{0.798} & \textbf{24.649} & \textbf{0.892} \\
      \hdashline
      GridDehazeNet-P2   & 16.826 & 0.763 & 18.854 & 0.842 & 15.632 & 0.784 & 15.279 & 0.787 & 15.828 & 0.613 & 19.264 & 0.695 & 9.332  & 0.585 \\
      GridDehazeNet-P4   & 13.636 & 0.634 & 16.447 & 0.756 & 12.155 & 0.660 & 14.300 & 0.718 & 15.713 & 0.505 & 18.597 & 0.565 & 8.067  & 0.459 \\
      GridDehazeNet-P6   & 12.342 & 0.561 & 15.269 & 0.688 & 10.504 & 0.575 & 13.521 & 0.654 & 15.536 & 0.414 & 17.714 & 0.459 & 7.268  & 0.389 \\

      \hdashline
      GridDehazeNet-P8   & \textcolor{red}{11.718} & \textcolor{red}{0.512} & \textcolor{red}{14.420} & \textcolor{red}{0.628} & \textcolor{red}{9.694}  & \textcolor{red}{0.517} & \textcolor{red}{12.932} & \textcolor{red}{0.597} & \textcolor{red}{15.319} & \textcolor{red}{0.348} & \textcolor{red}{16.838} & \textcolor{red}{0.380} & \textcolor{red}{6.753}  & \textcolor{red}{0.356} \\
      GridDehazeNet-M8   & \textcolor{blue}{13.064} & 0.635 & 16.758 & 0.766 & \textcolor{blue}{12.348} & \textcolor{blue}{0.700} & 14.150 & \textcolor{blue}{0.774} & 15.521 & 0.465 & 17.771 & 0.478 & 7.698  & 0.509 \\
      GridDehazeNet-I8   & 12.831 & \textcolor{blue}{0.656} & \textcolor{blue}{16.928} & \textcolor{blue}{0.778} & 12.280 & 0.658 & \textcolor{blue}{14.906} & 0.730 & \textcolor{blue}{15.933} & \textcolor{blue}{0.676} & \textcolor{blue}{19.800} & \textcolor{blue}{0.711} & \textcolor{blue}{10.083} & \textcolor{blue}{0.540} \\

    \hline
    \hline
      MSBDN              & \textbf{24.229} & \textbf{0.903} & \textbf{24.229} & \textbf{0.903} & \textbf{23.785} & \textbf{0.904} & \textbf{30.382} & \textbf{0.969} & \textbf{17.074} & \textbf{0.505} & \textbf{20.450} & \textbf{0.579} & \textbf{23.093} & \textbf{0.884} \\
      \hdashline
      MSBDN-P2           & 13.845 & 0.691 & 13.845 & 0.691 & 13.769 & 0.737 & 15.824 & 0.788 & 16.209 & 0.448 & 18.596 & 0.540 & 9.510  & 0.561 \\
      MSBDN-P4           & 10.098 & 0.531 & 10.098 & 0.531 & 9.746  & 0.605 & 12.874 & 0.643 & 14.914 & 0.352 & 16.422 & 0.468 & 7.422  & 0.401 \\
      MSBDN-P6           & 8.243  & 0.430 & 8.243  & 0.430 & 8.543  & 0.538 & 11.814 & 0.555 & 13.936 & 0.273 & 14.866 & 0.396 & 6.594  & 0.333 \\

      \hdashline
      MSBDN-P8           & \textcolor{red}{7.016}  & \textcolor{red}{0.361} & \textcolor{red}{7.016}  & \textcolor{red}{0.361} & \textcolor{red}{8.063}  & \textcolor{red}{0.496} & \textcolor{red}{11.148} & \textcolor{red}{0.491} & \textcolor{red}{13.237} & \textcolor{red}{0.219} & \textcolor{red}{13.985} & \textcolor{red}{0.346} & \textcolor{red}{6.215}  & \textcolor{red}{0.296} \\
      MSBDN-M8           & 8.948  & 0.488 & 15.347 & 0.703 & 9.550  & 0.602 & 13.025 & \textcolor{blue}{0.739} & 14.597 & 0.338 & 15.086 & 0.400 & 7.610  & 0.476 \\
      MSBDN-I8           & \textcolor{blue}{12.694} & \textcolor{blue}{0.622} & \textcolor{blue}{16.641} & \textcolor{blue}{0.767} & \textcolor{blue}{12.257} & \textcolor{blue}{0.644} & \textcolor{blue}{14.889} & 0.728 & \textcolor{blue}{15.496} & \textcolor{blue}{0.387} & \textcolor{blue}{16.766} & \textcolor{blue}{0.461} & \textcolor{blue}{9.993}  & \textcolor{blue}{0.573} \\

  \hline
  \end{tabular}
\end{table*}
%------------------------------------------------------------------------
%-------------------------------------------------------------------------
%-------------------------------------------------------------------------
%-------------------------------------------------------------------------
\section{Experiments}
The experiments contain three parts. Subsections \ref{subsec:experiments_attack} and \ref{subsec:experiments_defense} give the attack results and defense results, respectively. Subsection \ref{subsec:ablation_study_and_discussions} provides the ablation study and discussions. The P, M, I, G and S represent the $L_{P}^{MSE}$, $L_{M}^{MSE}$, $L_{I}^{MSE}$, $L_{G}^{MSE}$ and $L_{P}^{SSIM}$, respectively.
%------------------------------------------------------------------------
\subsection{Experiments on Attack}
\label{subsec:experiments_attack}

%-------------------------------------------------------------------------
%------------------------------------------------------------------------
\begin{table*}
  \scriptsize
  \centering
  \caption{Quantitative attack results obtained by $A_{N}$ and $L_{I}^{MSE}$. The worst attack result in each dehazing method is in \textcolor{blue}{blue}.}
  \label{tab:attack_results_epsilon8_noise_and_haze-preserved}
  \begin{tabular}{c|c@{\hspace{0.27cm}}c|c@{\hspace{0.27cm}}c|c@{\hspace{0.27cm}}c|c@{\hspace{0.27cm}}c|c@{\hspace{0.27cm}}c|c@{\hspace{0.27cm}}c|c@{\hspace{0.27cm}}c}
    \hline
      \multirow{2}{*}{Methods} & \multicolumn{2}{c|}{ITS} & \multicolumn{2}{c|}{OTS} & \multicolumn{2}{c|}{4K} & \multicolumn{2}{c|}{Foggy-City} & \multicolumn{2}{c|}{I-HAZE} & \multicolumn{2}{c|}{O-HAZE} & \multicolumn{2}{c}{D-Hazy}\\
    \cline{2-15}
     & PSNR$\uparrow$ & SSIM$\uparrow$ & PSNR$\uparrow$ & SSIM$\uparrow$ & PSNR$\uparrow$ & SSIM$\uparrow$ & PSNR$\uparrow$ & SSIM$\uparrow$ & PSNR$\uparrow$ & SSIM$\uparrow$ & PSNR$\uparrow$ & SSIM$\uparrow$ & PSNR$\uparrow$ & SSIM$\uparrow$ \\

    \hline                 
      4KDehazing-$A_{N}$      & \textcolor{blue}{23.407} & \textcolor{blue}{0.776} & \textcolor{blue}{26.25}1 & \textcolor{blue}{0.772} & \textcolor{blue}{20.378} & \textcolor{blue}{0.754} & \textcolor{blue}{21.061} & \textcolor{blue}{0.768} & \textcolor{blue}{16.737} & 0.582 & \textcolor{blue}{20.736} & \textcolor{blue}{0.729} & \textcolor{blue}{19.327} & \textcolor{blue}{0.695} \\
      4KDehazing-I      & 12.725 & 0.648 & 16.470 & 0.761 & 12.001 & 0.645 & 14.900 & 0.709 & 15.834 & \textcolor{blue}{0.587} & 17.274 & 0.695 & 9.985  & 0.568 \\

      \hline
      \hline

      DM2FNet-$A_{N}$         & \textcolor{blue}{24.479} & \textcolor{blue}{0.785} & \textcolor{blue}{27.437} & \textcolor{blue}{0.783} & \textcolor{blue}{22.250} & \textcolor{blue}{0.720} & \textcolor{blue}{25.088} & \textcolor{blue}{0.766} & \textcolor{blue}{16.921} & \textcolor{blue}{0.625} & \textcolor{blue}{23.200} & \textcolor{blue}{0.685} & \textcolor{blue}{18.877} & \textcolor{blue}{0.594} \\
      DM2FNet-I         & 12.207 & 0.644 & 16.285 & 0.748 & 12.367 & 0.634 & 14.852 & 0.754 & 14.681 & 0.558 & 15.541 & 0.640 & 9.729  & 0.582 \\

    \hline
    \hline

      FFANet-$A_{N}$          & \textcolor{blue}{23.930} & \textcolor{blue}{0.754} & \textcolor{blue}{26.790} & \textcolor{blue}{0.805} & \textcolor{blue}{19.497} & \textcolor{blue}{0.712} & \textcolor{blue}{16.832} & 0.699 & \textcolor{blue}{16.398} & 0.633 & \textcolor{blue}{22.085} & \textcolor{blue}{0.767} & \textcolor{blue}{16.028} & \textcolor{blue}{0.544} \\
      FFANet-I          & 12.386 & 0.653 & 16.608 & 0.771 & 12.805 & 0.613 & 14.899 & \textcolor{blue}{0.718} & 15.690 & \textcolor{blue}{0.688} & 19.412 & 0.717 & 10.338 & 0.528 \\
     
    \hline
    \hline
   
      GCANet-$A_{N}$          & \textcolor{blue}{25.166} & \textcolor{blue}{0.833} & \textcolor{blue}{26.275} & \textcolor{blue}{0.837} & \textcolor{blue}{22.037} & \textcolor{blue}{0.778} & \textcolor{blue}{25.270} & \textcolor{blue}{0.855} & \textcolor{blue}{16.441} & \textcolor{blue}{0.644} & \textcolor{blue}{20.222} & \textcolor{blue}{0.694} & \textcolor{blue}{17.706} & \textcolor{blue}{0.717} \\
      GCANet-I          & 13.433 & 0.645 & 17.739 & 0.755 & 12.010 & 0.599 & 15.656 & 0.726 & 15.843 & 0.585 & 17.659 & 0.600 & 10.492 & 0.563 \\

    \hline
    \hline

      GridDehazeNet-$A_{N}$   & \textcolor{blue}{22.304} & \textcolor{blue}{0.792} & \textcolor{blue}{25.974} & \textcolor{blue}{0.802} & \textcolor{blue}{20.384} & \textcolor{blue}{0.754} & \textcolor{blue}{17.223} & \textcolor{blue}{0.735} & 15.832 & 0.622 & 19.350 & \textcolor{blue}{0.722} & \textcolor{blue}{10.531} & \textcolor{blue}{0.583} \\
      GridDehazeNet-I   & 12.831 & 0.656 & 16.928 & 0.778 & 12.280 & 0.658 & 14.906 & 0.730 & \textcolor{blue}{15.933} & \textcolor{blue}{0.676} & \textcolor{blue}{19.800} & 0.711 & 10.083 & 0.540 \\

    \hline
    \hline

      MSBDN-$A_{N}$           & \textcolor{blue}{23.367} & \textcolor{blue}{0.787} & \textcolor{blue}{27.145} & \textcolor{blue}{0.811} & \textcolor{blue}{22.837} & \textcolor{blue}{0.792} & \textcolor{blue}{26.072} & \textcolor{blue}{0.776} & \textcolor{blue}{16.906} & \textcolor{blue}{0.414} & \textcolor{blue}{20.110} & \textcolor{blue}{0.513} & \textcolor{blue}{21.366} & \textcolor{blue}{0.694} \\
      MSBDN-I           & 12.694 & 0.622 & 16.641 & 0.767 & 12.257 & 0.644 & 14.889 & 0.728 & 15.496 & 0.387 & 16.766 & 0.461 & 9.993  & 0.573 \\

  \hline
  \end{tabular}
\end{table*}
%------------------------------------------------------------------------
%-------------------------------------------------------------------------

%-------------------------------------------------------------------------
%------------------------------------------------------------------------
\begin{figure*}
  \centering
  \includegraphics[width=18cm,height=2.8cm]{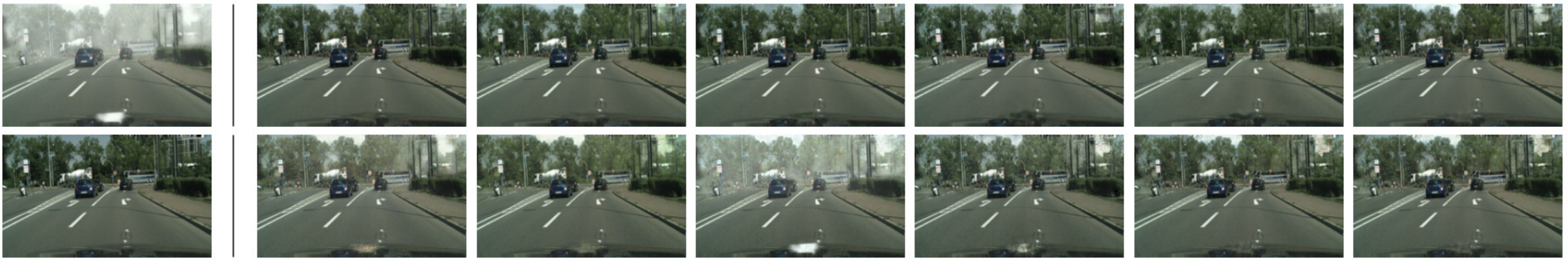}
 
  \leftline{\hspace{0.5cm} (a) {$I \& J$} \hspace{1.0cm} (b) 4KDehazing \hspace{0.3cm}  (c) DM2FNet \hspace{0.4cm}  (d) FFANet \hspace{0.5cm} (e) GCANet \hspace{0.1cm} (f) GridDehazeNet \hspace{0.05cm} (g) MSBDN}
  \caption{Dehazing visual results obtained by $A_{N}$. The $\epsilon$ of top row and bottom row are 0 and 8, respectively. Images are from Foggy-City.}
  \label{fig:noise_cityscapes}
\end{figure*}
%-------------------------------------------------------------------------
%------------------------------------------------------------------------

%------------------------------------------------------------------------
\begin{figure*}
  \centering
  \includegraphics[width=18cm,height=2.8cm]{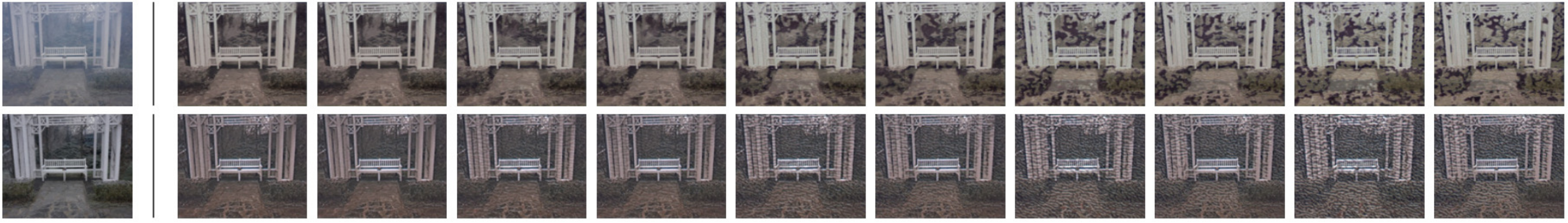}

  \leftline{\hspace{0.1cm} (a) {$I \& J$} \hspace{0.5cm} (b) P-0 \hspace{0.4cm} (c) M-0 \hspace{0.3cm} (d) P-2 \hspace{0.3cm} (e) M-2 \hspace{0.3cm} (f) P-4 \hspace{0.3cm} (g) M-4 \hspace{0.3cm} (h) P-6 \hspace{0.4cm} (i) M-6 \hspace{0.4cm} (j) P-8 \hspace{0.3cm} (k) M-8 }

  \caption{Attack visual results obtained by $L_{P}^{MSE}$ and $L_{M}^{MSE}$. The $P-\epsilon$ and $M-\epsilon$ denote the results when $\epsilon \in \{0, 2, 4, 6, 8\}$. The algorithms of each row from up to down are 4KDehazing and GridDehazeNet, respectively. Images are from O-HAZE.}
  \label{fig:OHAZE_mask_nomask}
\end{figure*}
%-------------------------------------------------------------------------

%-------------------------------------------------------------------------
%------------------------------------------------------------------------
\subsubsection{Settings on Attack}
\label{subsubsec:attack_settings}
To show the attack results on different datasets with different scales, the experiments are conducted on six different datasets: ITS (indoor) and OTS (outdoor) from RESIDE~\cite{li2018benchmarking}, I-HAZE~\cite{ancuti2018ihaze}, O-HAZE~\cite{ancuti2018ohaze}, D-Hazy~\cite{ancuti2016d}, 4K~\cite{zheng2021ultra} and Foggy-City~\cite{sakaridis2018semantic,cordts2016cityscapes}. The details about these datasets can be found in corresponding papers.

To ensure that the attack results are reliable, we choose the widely cited papers (dehazing algorithms) that are published on well-known conferences as baselines. The baselines methods are DM2FNet~\cite{deng2019deep}, GCANet~\cite{chen2019gated}, MSBDN~\cite{dong2020multi}, FFANet~\cite{qin2020ffa}, GridDehazeNet~\cite{liu2019griddehazenet} and 4KDehazing~\cite{zheng2021ultra}, respectively. The batch size is the same for all the baseline dehazing networks trained on the same dataset, that is $8$ for ITS/OTS/4K and $4$ for D-Hazy/I-HAZE/O-HAZE/Foggy-City. The image size for ITS/OTS/4K/I-HAZE/O-HAZE, D-Hazy and Foggy-City are $256 \times 256$,  $480 \times 640$ and $256 \times 512$, respectively. We choose the widely used metric PSNR and SSIM~\cite{wang2004image} for the evaluation, which are consistent with our attack loss functions. Lower PSNR and SSIM value stand for worse dehazing result, while representing better attack performance. When evaluating the performance of the model before and after the attack, we used exactly the same experimental settings. The dimensionalities of the input hazy image and the parameters of the dehazing model are exactly the same. Therefore, the reason for the degradation in model performance is the adversarial perturbations generated by our proposed attack algorithm.

The choice of parameters is theoretically infinite. For the attack experiments, the $\epsilon$ is chosen from $\{0, 2, 4, 6, 8\}$, and $\alpha$ is set to $2$. Both $\epsilon$ and $\alpha$ are divided by 255. The $\epsilon = 0$ means no attack. The attack iteration steps are $10$. Initialized values of $\delta$ for attacking and the noise for noise attack $A_{N}$ are both sampled from uniform distribution $U(-\epsilon, \epsilon)$. The attack methods contain $L_{P}^{MSE}$, $A_{N}$, $L_{M}^{MSE}$, $L_{G}^{MSE}$, $L_{I}^{MSE}$ and $L_{P}^{SSIM}$, respectively. The attack process is to maximize these loss functions (except noise attack $A_{N}$). It is worth noting that the $mask$ in $L_{M}$ ($L_{M}^{MSE}$) has three channels obtained by copying a single channel. The $\mu$ used to obtain the $mask$ is calculated on the whole image (channel average) as shown in Equation \ref{eq:mask_attack}. Any position $(m, n)$ in the hazy image $I$ is either be attacked or not attacked. Therefore, $mask(m, n)$ equals $1$ when any channel in position $(m, n)$ satisfied $I(m, n) - J_{p}(m, n) > \mu$. 
%-------------------------------------------------------------------------
%------------------------------------------------------------------------

%------------------------------------------------------------------------
\begin{figure*}
  \centering
  \includegraphics[width=18cm,height=2.8cm]{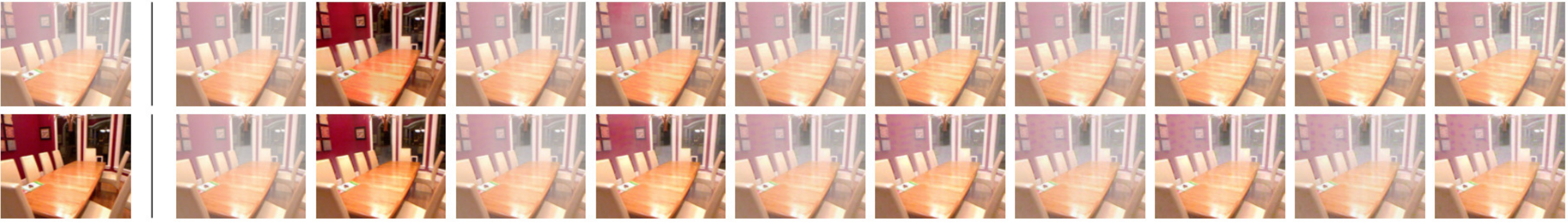}
 
  \leftline{\hspace{0.05cm} (a) {$I \& J$} \hspace{0.6cm} (b) $I^{\delta}$-0 \hspace{0.15cm} (c)$J_{p}^{\delta}$-0 \hspace{0.3cm} (d) $I^{\delta}$-2 \hspace{0.2cm} (e) $J_{p}^{\delta}$-2 \hspace{0.2cm} (f) $I^{\delta}$-4 \hspace{0.25cm} (g) $J_{p}^{\delta}$-4 \hspace{0.15cm} (h) $I^{\delta}$-6 \hspace{0.25cm} (i) $J_{p}^{\delta}$-6 \hspace{0.25cm} (j) $I^{\delta}$-8 \hspace{0.25cm} (k) $J_{p}^{\delta}$-8}

  \caption{Attack visual results obtained by $L_{I}^{MSE}$. The attacked hazy image and corresponding prediction (dehazed image) are denoted as $I^{\delta}-\epsilon$ and $J_{p}^{\delta}-\epsilon$, respectively. $\epsilon \in {\{0, 2, 4, 6, 8\}}$. The algorithms of each row from up to down are 4KDehazing and GridDehazeNet, respectively. Images are from ITS.}
  \label{fig:visual_ITS_input}
\end{figure*}
%-------------------------------------------------------------------------

%------------------------------------------------------------------------
\begin{figure*}
  \centering
  \includegraphics[width=18cm,height=2.8cm]{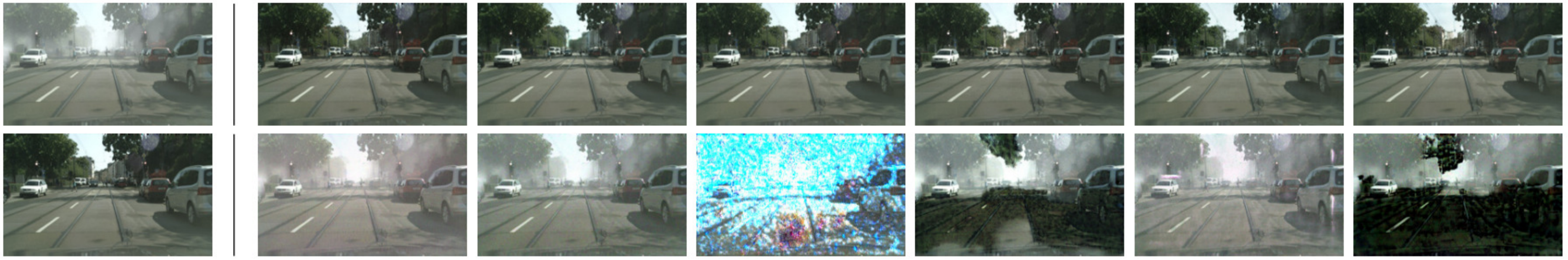}

  \leftline{\hspace{0.5cm} (a) {$I \& J$} \hspace{0.9cm} (b) 4KDehazing \hspace{0.2cm}   (c) DM2FNet \hspace{0.5cm}  (d) FFANet \hspace{0.4cm} (e) GCANet \hspace{0.20cm} (f) GridDehazeNet \hspace{0.05cm} (g) MSBDN}
  \caption{Attack visual results obtained by attack loss $L_{G}^{MSE}$. The $\epsilon$ of each row from up to down are 0 and 8, respectively. Images are from Foggy-City.}
  \label{fig:cityscapes_gt}
\end{figure*}
%-------------------------------------------------------------------------

%------------------------------------------------------------------------
\begin{figure*}
  \centering
  \includegraphics[width=18cm,height=2.8cm]{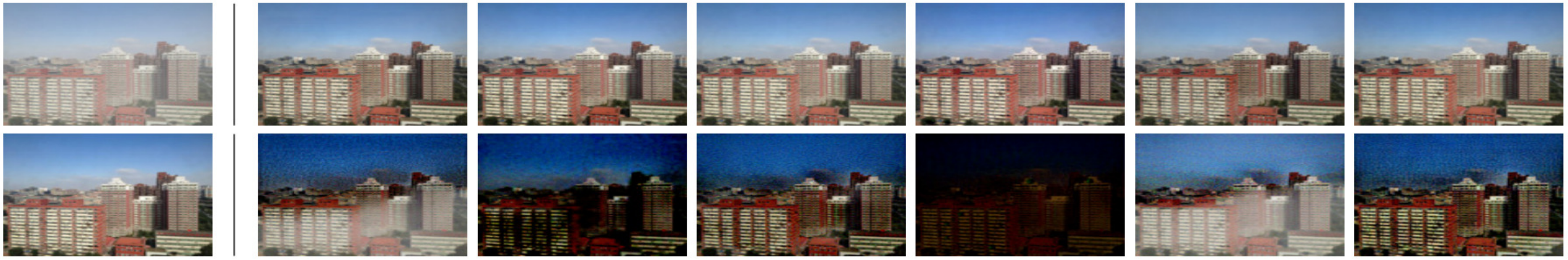}
 
  \leftline{\hspace{0.5cm} (a) {$I \& J$} \hspace{0.9cm} (b) 4KDehazing \hspace{0.2cm}   (c) DM2FNet \hspace{0.5cm}  (d) FFANet \hspace{0.4cm} (e) GCANet \hspace{0.20cm} (f) GridDehazeNet \hspace{0.05cm} (g) MSBDN}
  \caption{Image dehazing visual results obtained by different dehazing algorithms under attack loss $L_{P}^{SSIM}$. The $\epsilon$ of each row from up to down are 0 and 8, respectively. Images are from OTS.}
  \label{fig:OTS_SSIM}
\end{figure*}
%------------------------------------------------------------------------

% ---------------------------------------------------------------------------------------------------------------
\begin{figure*}
  \centering
  \includegraphics[width=18cm,height=2.8cm]{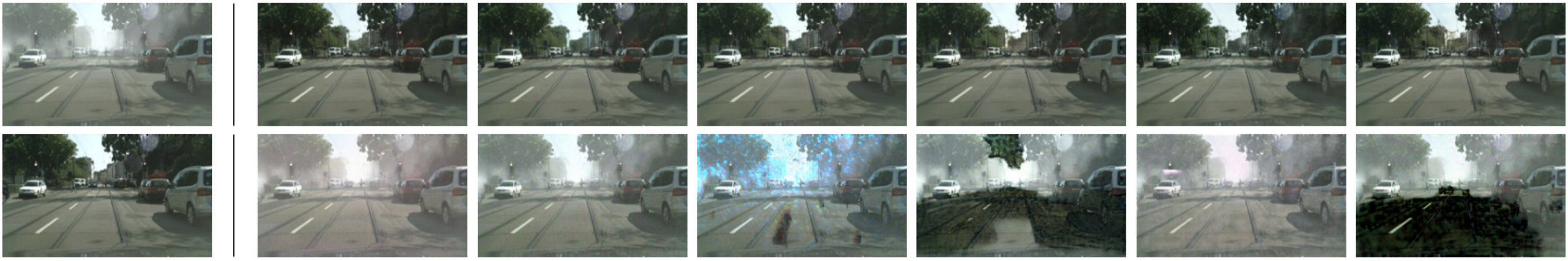}

  \leftline{\hspace{0.5cm} (a) {$I \& J$} \hspace{0.9cm} (b) 4KDehazing \hspace{0.2cm}   (c) DM2FNet \hspace{0.5cm}  (d) FFANet \hspace{0.4cm} (e) GCANet \hspace{0.20cm} (f) GridDehazeNet \hspace{0.05cm} (g) MSBDN}
  \caption{Image dehazing visual results obtained by different dehazing algorithms under attack loss $L_{P}^{MSE}$. The $\epsilon$ of each row from up to down are 0 and 8, respectively. Images are from Foggy-City.}
  \label{fig:pred_mse_cityscapes}
\end{figure*}
% ---------------------------------------------------------------------------------------------------------------

%------------------------------------------------------------------------
%-------------------------------------------------------------------------

\begin{figure}
  \centering
  \footnotesize
  \includegraphics[width=9cm,height=13.6cm]{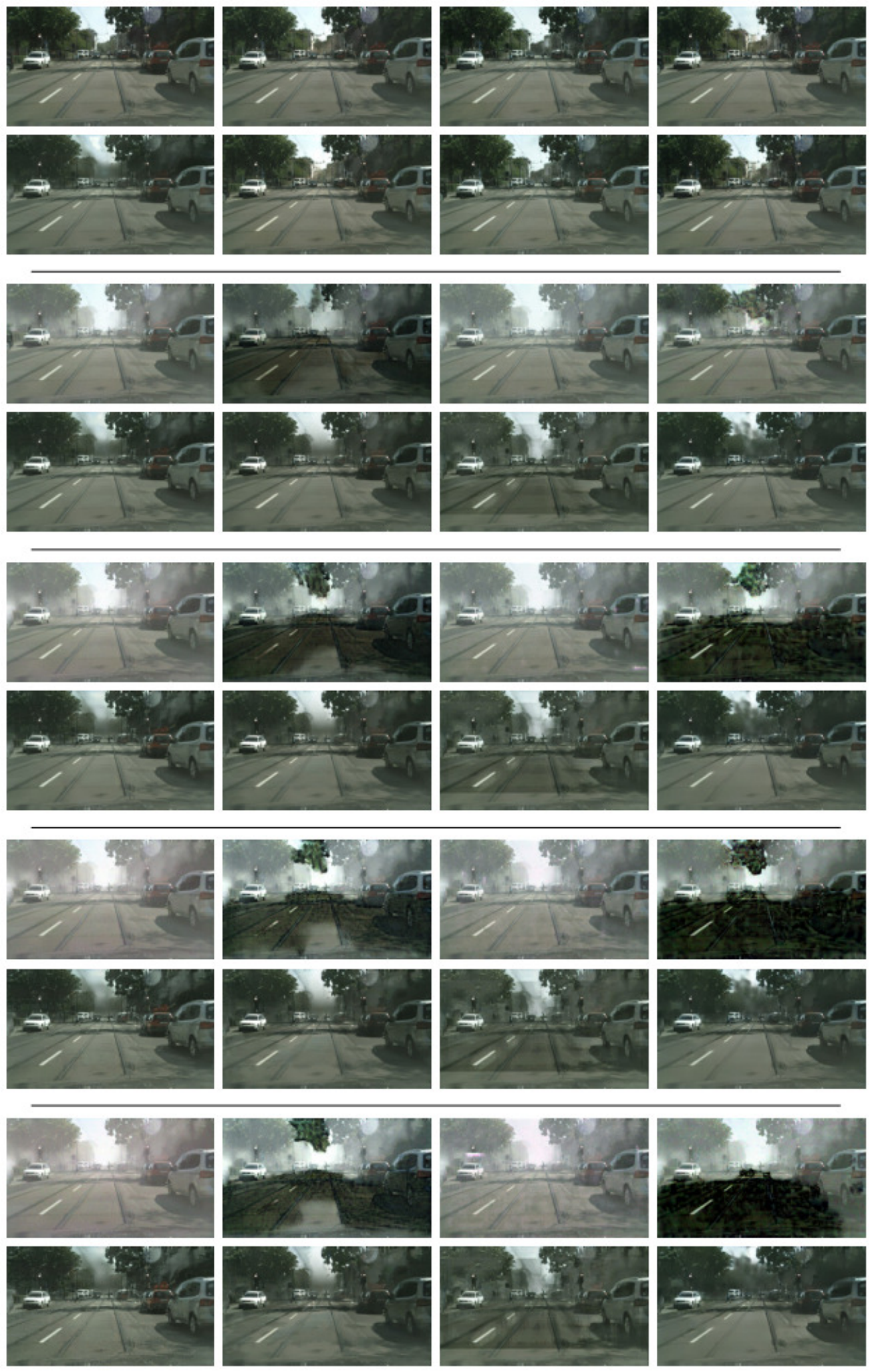}

  \leftline{\hspace{0.1cm} (a) 4KDehazing \hspace{0.4cm}  (b) GCANet \hspace{0.3cm} (c) GridDehazeNet \hspace{0.2cm} (d) MSBDN}
  \caption{Image dehazing results obtained by different dehazing algorithms under attack loss $L_{P}^{MSE}$ before and after defense training. Every two rows of images are in a group. The above images of each group are the results of direct attack (without defense training), and the bottom images are the results of attack after defense training. The $\epsilon$ of two rows from up to down are 0, 2, 4, 6 and 8, respectively. Images are from Foggy-City.}
  \label{fig:pred_mse_cityscapes_before_and_after_defense}
\end{figure}
%-------------------------------------------------------------------------
%------------------------------------------------------------------------

%-------------------------------------------------------------------------
%------------------------------------------------------------------------

\subsubsection{Results on Attack}
Here, quantitative and visual evaluations of the attack results are given. Furthermore, the corresponding conclusions and analyses are provided.
\begin{itemize}
\item Attack by $L_{P}^{MSE}$: Table \ref{tab:attack_results_differ_epsilon_by_LP_LM_LI} shows the dehazing performance obtained by different dehazing methods on various benchmark datasets without being attacked. When there is no attack, these dehazing models are able to achieve impressive performance. However, when they were attacked by our proposed $L_{P}^{MSE}$, the values of PSNR and SSIM dropped significantly. As $\epsilon$ increases from 2 to 8, the attack performance gradually increases.  The visual results obtained by the dehazing networks after attacks on the Foggy-City dataset are shown in Figure \ref{fig:pred_mse_cityscapes}. It can be seen that the visual dehazing results after the attack are quite unsatisfactory. The quantitative and visual results demonstrate that the attacker can use predicted dehazed images obtained by dehazing networks as pseudo labels to attack the dehazing networks.
% ---------------------------------------------------------------------------------------------------------------
\item Attack by $L_{M}^{MSE}$: Table \ref{tab:attack_results_differ_epsilon_by_LP_LM_LI} shows the attack results obtained by $L_{M}^{MSE}$. Overall, the attack results obtained by $L_{M}^{MSE}$ are slightly lower than those of $L_{P}^{MSE}$. This result is intuitive since $L_{M}^{MSE}$ only performed the attack on areas with high haze density, rather than the entire hazy image. Figure \ref{fig:OHAZE_mask_nomask} compares the visual performance between the haze layer mask attack and the non-mask attack method. The haze layer mask attack can achieve similar effects as the non-mask attack, which shows that the attacker can partially attack the input hazy image, rather than the whole image.
% ---------------------------------------------------------------------------------------------------------------
\item Attack by $L_{I}^{MSE}$: As shown in Table \ref{tab:attack_results_differ_epsilon_by_LP_LM_LI}, the attack results obtained by $L_{MSE}$ are lower than those of $L_{P}^{MSE}$ and $L_{M}^{MSE}$. The possible reason for this phenomenon is that the purpose of $L_{I}$ is to preserve the haze in the image, which does not destroy the content and structure of the image. The attacked hazy images $I^{\delta} = I + \delta$ and predicted attacked dehazed images $J_{p}^{\delta}$ shown in Figure \ref{fig:visual_ITS_input} are quite close from the perspective of visual perception when $\epsilon > 0$. This phenomenon illustrates that the haze-preserved attack loss $L_{I}^{MSE}$ can preserve the haze inside the hazy images.
% ---------------------------------------------------------------------------------------------------------------
\item Attack by $L_{P}^{SSIM}$: Table \ref{tab:attack_results_MSE_SSIM} compares the values of quantitative results obtained by $L_{P}^{SSIM}$ and $L_{P}^{MSE}$. The $\epsilon$ is set to $8$. $L_{P}^{SSIM}$ is a direct attack on SSIM metrics. $L_{P}^{MSE}$ is an approximately direct attack on PSNR since PSNR is inversely proportional to MSE. The results show that the value of PSNR decreases more when using $L_{P}^{MSE}$, while the value of SSIM decreases more when using $L_{P}^{SSIM}$. Quantitative attack results show that the first-order gradient can effectively attack both PSNR and SSIM metrics.  Figure \ref{fig:OTS_SSIM} shows the outdoor attack results obtained by attack loss $L_{P}^{SSIM}$. Under the attack of $L_{P}^{SSIM}$, the visual quality of dehazed images all decreases, which shows the effectiveness of the attack method based on SSIM loss.
\end{itemize}

According to the experimental results obtained by attack methods $L_{P}^{MSE}$, $L_{M}^{MSE}$, $L_{I}^{MSE}$ and $L_{P}^{SSIM}$, three valuable conclusions can be drawn. First, the attacker can use first-order gradient information to successfully attack the dehazing model. Second, using dehazed images as attack targets achieves better attack results than using hazy images. Third, taking $MSE$ and $SSIM$ as the loss function in the attack process can improve the attack performance of PSNR and SSIM, respectively.
%-------------------------------------------------------------------------
%------------------------------------------------------------------------

%-------------------------------------------------------------------------
%------------------------------------------------------------------------
\begin{table*}
  \scriptsize
  \centering
  \caption{Quantitative attack results obtained by $L_{P}^{MSE}$ and $L_{G}^{MSE}$. The best attack result in each dehazing method is in \textcolor{red}{red}.}
  \label{tab:attack_results_epsilon8_P_and_G}
  \begin{tabular}{c|c@{\hspace{0.27cm}}c|c@{\hspace{0.27cm}}c|c@{\hspace{0.27cm}}c|c@{\hspace{0.27cm}}c|c@{\hspace{0.27cm}}c|c@{\hspace{0.27cm}}c|c@{\hspace{0.27cm}}c}
    \hline
      \multirow{2}{*}{Methods} & \multicolumn{2}{c|}{ITS} & \multicolumn{2}{c|}{OTS} & \multicolumn{2}{c|}{4K} & \multicolumn{2}{c|}{Foggy-City} & \multicolumn{2}{c|}{I-HAZE} & \multicolumn{2}{c|}{O-HAZE} & \multicolumn{2}{c}{D-Hazy}\\
    \cline{2-15}
     & PSNR$\uparrow$ & SSIM$\uparrow$ & PSNR$\uparrow$ & SSIM$\uparrow$ & PSNR$\uparrow$ & SSIM$\uparrow$ & PSNR$\uparrow$ & SSIM$\uparrow$ & PSNR$\uparrow$ & SSIM$\uparrow$ & PSNR$\uparrow$ & SSIM$\uparrow$ & PSNR$\uparrow$ & SSIM$\uparrow$ \\

    \hline                 
      4KDehazing-P      & 10.545 & 0.533 & 13.181 & 0.605 & \textcolor{red}{6.617}  & \textcolor{red}{0.315} & 12.355 & 0.634 & 15.235 & 0.426 & 16.744 & \textcolor{red}{0.533} & 8.061  & 0.426 \\
      4KDehazing-G      & \textcolor{red}{10.131} & \textcolor{red}{0.505} & \textcolor{red}{11.845} & \textcolor{red}{0.532} & 7.031  & 0.359 & \textcolor{red}{12.250} & \textcolor{red}{0.625} & \textcolor{red}{13.335} & \textcolor{red}{0.394} & \textcolor{red}{14.950} & 0.537 & \textcolor{red}{6.713}  & \textcolor{red}{0.278} \\

      \hline
      \hline

      DM2FNet-P         & 11.214 & 0.570 & 14.464 & 0.617 & 8.624  & 0.476 & 14.010 & 0.650 & 13.265 & 0.495 & 10.899 & 0.454 & 6.813  & 0.268 \\
      DM2FNet-G         & \textcolor{red}{11.073} & \textcolor{red}{0.556} & \textcolor{red}{10.479} & \textcolor{red}{0.371} & \textcolor{red}{7.933}  & \textcolor{red}{0.427} & \textcolor{red}{13.763} & \textcolor{red}{0.642} & \textcolor{red}{11.893} & \textcolor{red}{0.442} & \textcolor{red}{10.785} & \textcolor{red}{0.421} & \textcolor{red}{6.537}  & \textcolor{red}{0.235} \\

    \hline
    \hline

      FFANet-P          & 10.091 & 0.448 & 13.978 & 0.611 & 8.742  & \textcolor{red}{0.441} & \textcolor{red}{9.728}  & \textcolor{red}{0.326} & 15.664 & \textcolor{red}{0.400} & 17.487 & \textcolor{red}{0.579} & 6.556  & \textcolor{red}{0.223} \\
      FFANet-G          & \textcolor{red}{10.063} & \textcolor{red}{0.437} & \textcolor{red}{12.279} & \textcolor{red}{0.537} & \textcolor{red}{8.726}  & 0.448 & 9.760  & \textcolor{red}{0.326} & \textcolor{red}{14.010} & 0.499 & \textcolor{red}{15.938} & 0.601 & \textcolor{red}{6.552}  & 0.227 \\

    \hline
    \hline

      GCANet-P          & 7.530  & \textcolor{red}{0.283} & 8.733  & 0.407 & 7.990  & \textcolor{red}{0.409} & \textcolor{red}{10.433} & \textcolor{red}{0.516} & 13.353 & 0.424 & 14.200 & \textcolor{red}{0.455} & 6.932  & 0.354 \\
      GCANet-G          & \textcolor{red}{7.111}  & 0.299 & \textcolor{red}{7.557}  & \textcolor{red}{0.272} & \textcolor{red}{7.789}  & 0.421 & 10.538 & 0.522 & \textcolor{red}{11.907} & \textcolor{red}{0.422} & \textcolor{red}{13.990} & 0.465 & \textcolor{red}{6.822} & \textcolor{red}{0.346} \\

    \hline
    \hline

      GridDehazeNet-P   & \textcolor{red}{11.718} & \textcolor{red}{0.512} & 14.420 & 0.628 & 9.694  & 0.517 & 12.932 & 0.597 & 15.319 & \textcolor{red}{0.348} & 16.838 & \textcolor{red}{0.380} & 6.753  & 0.356 \\
      GridDehazeNet-G   & 11.956 & 0.581 & \textcolor{red}{12.156} & \textcolor{red}{0.547} & \textcolor{red}{9.416}  & \textcolor{red}{0.491} & \textcolor{red}{12.700} & \textcolor{red}{0.586} & \textcolor{red}{14.011} & 0.434 & \textcolor{red}{14.950} & 0.429 & \textcolor{red}{6.714}  & \textcolor{red}{0.354} \\

    \hline
    \hline
      MSBDN-P           & 7.016  & \textcolor{red}{0.361} & \textcolor{red}{7.016}  & \textcolor{red}{0.361} & 8.063  & 0.496 & 11.148 & 0.491 & 13.237 & 0.219 & 13.985 & 0.346 & 6.215  & 0.296 \\
      MSBDN-G           & \textcolor{red}{6.875}  & 0.388 & 11.523 & 0.522 & \textcolor{red}{7.829}  & \textcolor{red}{0.457} & \textcolor{red}{10.713} & \textcolor{red}{0.442} & \textcolor{red}{12.105} & \textcolor{red}{0.165} & \textcolor{red}{12.420} & \textcolor{red}{0.298} & \textcolor{red}{5.738}  & \textcolor{red}{0.226} \\
      
  \hline
  \end{tabular}
\end{table*}
%------------------------------------------------------------------------
%-------------------------------------------------------------------------

%-------------------------------------------------------------------------
%------------------------------------------------------------------------
\begin{table*}
  \scriptsize
  \centering
  \caption{Quantitative attack results obtained by $L_{P}^{MSE}$ and $L_{P}^{SSIM}$. The best attack result in each dehazing method is in \textcolor{red}{red}.}
  \label{tab:attack_results_MSE_SSIM}
  \begin{tabular}{c|c@{\hspace{0.27cm}}c|c@{\hspace{0.27cm}}c|c@{\hspace{0.27cm}}c|c@{\hspace{0.27cm}}c|c@{\hspace{0.27cm}}c|c@{\hspace{0.27cm}}c|c@{\hspace{0.27cm}}c}
    \hline
      \multirow{2}{*}{Methods} & \multicolumn{2}{c|}{ITS} & \multicolumn{2}{c|}{OTS} & \multicolumn{2}{c|}{4K} & \multicolumn{2}{c|}{Foggy-City} & \multicolumn{2}{c|}{I-HAZE} & \multicolumn{2}{c|}{O-HAZE} & \multicolumn{2}{c}{D-Hazy}\\
    \cline{2-15}
     & PSNR$\uparrow$ & SSIM$\uparrow$ & PSNR$\uparrow$ & SSIM$\uparrow$ & PSNR$\uparrow$ & SSIM$\uparrow$ & PSNR$\uparrow$ & SSIM$\uparrow$ & PSNR$\uparrow$ & SSIM$\uparrow$ & PSNR$\uparrow$ & SSIM$\uparrow$ & PSNR$\uparrow$ & SSIM$\uparrow$ \\

    \hline                 
      4KDehazing-P      & \textcolor{red}{10.545} & 0.533 & \textcolor{red}{13.181} & 0.605 & \textcolor{red}{6.617}  & 0.315 & \textcolor{red}{12.355} & 0.634 & \textcolor{red}{15.235} & 0.426 & \textcolor{red}{16.744} & 0.533 & \textcolor{red}{8.061}  & 0.426 \\
      4KDehazing-S      & 12.381 & \textcolor{red}{0.357} & 13.436 & \textcolor{red}{0.424} & 10.547 & \textcolor{red}{0.193} & 13.273 & \textcolor{red}{0.433} & 15.589 & \textcolor{red}{0.254} & 17.733 & \textcolor{red}{0.259} & 9.135  & \textcolor{red}{0.101} \\
      \hline
      \hline

      DM2FNet-P         & \textcolor{red}{11.214} & 0.570 & 14.464 & 0.617 & \textcolor{red}{8.624} & 0.476 & \textcolor{red}{14.010} & 0.650 & \textcolor{red}{13.265} & 0.495 & \textcolor{red}{10.899} & \textcolor{red}{0.454} & \textcolor{red}{6.813}  & 0.268 \\
      DM2FNet-S         & 11.411 & \textcolor{red}{0.397} & \textcolor{red}{12.691} & \textcolor{red}{0.409} & 9.542  & \textcolor{red}{0.328} & 14.699 & \textcolor{red}{0.433} & 15.561 & \textcolor{red}{0.388} & 14.851 & 0.509 & 8.142  & \textcolor{red}{0.081} \\

    \hline
    \hline
      FFANet-P          & \textcolor{red}{10.091} & 0.448 & \textcolor{red}{13.978} & 0.611 & \textcolor{red}{8.742}  & 0.441 & 9.728  & 0.326 & 15.664 & 0.400 & \textcolor{red}{17.487} & 0.579 & \textcolor{red}{6.556}  & 0.223 \\
      FFANet-S          & 11.072 & \textcolor{red}{0.245} & 14.456 & \textcolor{red}{0.438} & 11.359 & \textcolor{red}{0.273} & \textcolor{red}{8.702}  & \textcolor{red}{0.187} & \textcolor{red}{15.554} & \textcolor{red}{0.283} & 19.492 & \textcolor{red}{0.343} & 9.615  & \textcolor{red}{0.022} \\
      
    \hline
    \hline
      GCANet-P          & \textcolor{red}{7.530}  & 0.283 & \textcolor{red}{8.733}  & 0.407 & \textcolor{red}{7.990}  & 0.409 & \textcolor{red}{10.433} & 0.516 & \textcolor{red}{13.353} & 0.424 & \textcolor{red}{14.200} & 0.455 & \textcolor{red}{6.932}  & 0.354 \\
      GCANet-S          & 7.904  & \textcolor{red}{0.198} & 10.084 & \textcolor{red}{0.334} & 10.742 & \textcolor{red}{0.153} & 12.866 & \textcolor{red}{0.342} & 14.834 & \textcolor{red}{0.219} & 16.660 & \textcolor{red}{0.208} & 8.989 & \textcolor{red}{0.185} \\

    \hline
    \hline
      GridDehazeNet-P   & \textcolor{red}{11.718} & 0.512 & 14.420 & 0.628 & \textcolor{red}{9.694}  & 0.517 & \textcolor{red}{12.932} & 0.597 & 15.319 & 0.348 & \textcolor{red}{16.838} & 0.380 & \textcolor{red}{6.753} & 0.356 \\
      GridDehazeNet-S   & 13.083 & \textcolor{red}{0.395} & \textcolor{red}{13.476} & \textcolor{red}{0.435} & 13.179 & \textcolor{red}{0.341} & 14.618 & \textcolor{red}{0.420} & \textcolor{red}{15.189} & \textcolor{red}{0.264} & 17.243 & \textcolor{red}{0.173} & 9.337  & \textcolor{red}{0.158} \\

    \hline
    \hline
      MSBDN-P           & \textcolor{red}{7.016}  & 0.361 & \textcolor{red}{7.016}  & \textcolor{red}{0.361} & \textcolor{red}{8.063}  & 0.496 & \textcolor{red}{11.148} & 0.491 & \textcolor{red}{13.237} & 0.219 & \textcolor{red}{13.985} & 0.346 & \textcolor{red}{6.215}  & 0.296 \\
      MSBDN-S           & 8.453  & \textcolor{red}{0.316} & 12.928 & 0.435 & 12.671 & \textcolor{red}{0.309} & 14.201 & \textcolor{red}{0.248} & 14.945 & \textcolor{red}{0.182} & 17.269 & \textcolor{red}{0.222} & 7.595  & \textcolor{red}{0.129} \\
      
  \hline
  \end{tabular}
\end{table*}
%------------------------------------------------------------------------
%-------------------------------------------------------------------------
\subsection{Experiments on Defense}
\label{subsec:experiments_defense}
%-------------------------------------------------------------------------
%------------------------------------------------------------------------
\subsubsection{Settings on Defense}
The $\lambda$ in Equations (\ref{eq:L_def_P}) and (\ref{eq:L_def_G}) are set to 1. The epochs for defense training are set to 40 and $\epsilon$ is fixed as $8$ (maximum attack degree). Since the adversarial examples are generated dynamically, the adversarial defense training time under each epoch is almost $k$ (iteration steps for generating $\delta$) times of the original training time. The defense training strategy is verified by using 4KDehazing, GCANet, GridDehazeNet and MSBDN.

%-------------------------------------------------------------------------
%------------------------------------------------------------------------

\subsubsection{Results on Defense}
\label{subsec:results_on_defense}
Here we show the attack results after adversarial defense training both quantitatively and visually. In addition, the summaries and analyses of the defense results are also provided. The ``Method-$\epsilon$'' denote the attack results when $\epsilon=i, i \in \{2, 4, 6, 8\}$. TP denotes trained by $L_{def}^{P}$, and TG means trained by $L_{def}^{G}$. Table \ref{tab:defense_results_adv_training} shows the quantitative attack results before and after defense training. The results summarized from Table \ref{tab:defense_results_adv_training} are as follows.
\begin{itemize} 
  \item The adversarial defense training is quite effective for protecting the dehazing networks and the dehazing performance after defense training is obviously better (higher PSNR and SSIM) than that before defense training. Meanwhile, with the increases of $\epsilon$, the PSNR and SSIM that after defense training decrease more slowly than that before defense training.
  \item ``TP'' and ``TG'' in Table \ref{tab:defense_results_adv_training} show the dehazing results on original hazy examples $I$ (without attack). It can be seen that adversarial defense training will slightly reduce the dehazing performance on original hazy examples in most cases.
\end{itemize}

Figure \ref{fig:pred_mse_cityscapes_before_and_after_defense} shows the visual attack results before and after defense training. It can be seen that the attack results after defense training are more visually pleasing than those before defense training. The quality of image texture and detail is improved by defense training.

The main conclusion summarized from the quantitative and visual results is that the adversarial defense training can protect the dehazing networks, but it can not ``totally'' eliminate the negative effects caused by the attacker. The reason is that the dehazing model has seen the adversarial examples during defense training, so its robustness is improved. However, the attacker is still able to obtain the gradient information of the dehazing model that has performed the defense training~\cite{akhtar2021advances}. Therefore, the attacker can still suboptimally find the weaknesses of the dehazing model. The distribution of original hazy images is denoted as $P_{I}$, the distribution of original hazy images and hazy images that after adding adversarial perturbations is denoted as $P_{I + I^{\delta}}$, and distribution of clear images $J$ is $P_{J}$. The regular training process is to learn the mapping from $P_{I}$ to $P_{J}$, while the defense training process learns the mapping from $P_{I + I^{\delta}}$ to $P_{J}$. The goals of defense training and regular training are different. This phenomenon is consistent with the classification task~\cite{madry2017towards}, which leads to the need for a balance between the robustness and the performance of the model~\cite{robustness2022pang}. Therefore, the performance of the dehazing model that after defense training on the original hazy image will be slightly worse.

%-------------------------------------------------------------------------
%------------------------------------------------------------------------

\subsubsection{Analysis of the Teacher Network}

Table \ref{tab:gt_defense_pred_mse_attack_compared_with_pred_mse_defense_pred_mse_attack} shows the results obtained using $L_{P}^{MSE}$ for attack after defense training using $L_{def}^{P}$ and $L_{def}^{G}$, respectively. The results in Table \ref{tab:gt_defense_pred_mse_attack_compared_with_pred_mse_defense_pred_mse_attack} are on the Foggy-City dataset. The results show that the defense results obtained by using $L_{def}^{G}$ and $L_{def}^{P}$ against $L_{P}^{MSE}$ are comparable. The results demonstrate that the additional computational cost of using a Teacher network can be avoided.

%-------------------------------------------------------------------------
%------------------------------------------------------------------------

\subsubsection{Multi-step Attack and Defense}
The attacker takes an iteration-based strategy to attack the dehazing networks. The iteration step is $k$. To investigate how $k$ affects the attack and defense performance, we set $k$ to different values and calculate the corresponding dehazing results. The $\epsilon$ is set to $4$. Table \ref{tab:defense_results_differ_attack_iterations} shows the attack results obtained by attack loss $L_{P}^{MSE}$ when $k \in \{1, 2, 3, 5, 10, 15, 20, 25, 30\}$. The results in Table \ref{tab:defense_results_differ_attack_iterations} indicate that with the increases of $k$, the attack performance increases in most cases. The attack effect does not increase significantly when $k > 10$. The possible reason for this phenomenon is that the value of $\epsilon$ is limited to $4$. Therefore, the attack effect of an attacker who modifies the pixel value within the range $(-4, 4)$ will not continue to grow rapidly.

For adversarial defense training stage, $k$ is randomly selected from $\{20, 25, 30\}$ in each iteration. The attack results before and after defense training are shown in Table \ref{tab:defense_results_differ_attack_iterations}, which can achieve the same conclusions as the situation where the step size of defense training is fixed. The results indicate that adversarial defense training with the multi-step attack is effective when the attacker adopts different attack iterations ($k$) in most cases.

%-------------------------------------------------------------------------
%------------------------------------------------------------------------

%-------------------------------------------------------------------------
%------------------------------------------------------------------------
\begin{table*}
  \footnotesize
  \centering
  \caption{Attack result obtained by $L_{P}^{MSE}$ after training by $L_{def}^{G}$ and $L_{def}^{P}$, respectively. PSNR and SSIM are displayed in the form of A/B, where A denotes the defense loss $L_{def}^{G}$ and B means the defense loss $L_{def}^{P}$.}
  \label{tab:gt_defense_pred_mse_attack_compared_with_pred_mse_defense_pred_mse_attack}
  \begin{tabular}{c|c@{\hspace{0.8cm}}c|c@{\hspace{0.8cm}}c|c@{\hspace{0.5cm}}c|c@{\hspace{0.8cm}}c}
    \hline
      \multirow{2}{*}{Methods} & \multicolumn{2}{c|}{4KDehazing} & \multicolumn{2}{c|}{GCANet} & \multicolumn{2}{c|}{GridDehazeNet} & \multicolumn{2}{c}{MSBDN}\\
    \cline{2-9}
     & PSNR$\uparrow$ & SSIM$\uparrow$ & PSNR$\uparrow$ & SSIM$\uparrow$ & PSNR$\uparrow$ & SSIM$\uparrow$ & PSNR$\uparrow$ & SSIM$\uparrow$ \\
    
     \hline
      $\epsilon = 2$           & 24.258/\textbf{24.413} & 0.905/\textbf{0.906} & \textbf{23.619}/23.486 & \textbf{0.899}/0.895 
                  & 20.763/\textbf{21.059} & \textbf{0.858}/0.850 & \textbf{23.545}/23.380 & \textbf{0.891}/0.887 \\

      $\epsilon = 4$           & 23.604/\textbf{23.918} & \textbf{0.885}/0.884 & \textbf{22.701}/22.581 & \textbf{0.875}/0.872
                  & 20.175/\textbf{20.406} & \textbf{0.832}/0.819 & \textbf{22.316}/22.211 & \textbf{0.868}/0.865 \\

      $\epsilon = 6$           & 22.843/\textbf{23.193} & \textbf{0.853}/0.850 & \textbf{21.814}/21.701 & \textbf{0.844}/0.842
                  & 19.631/\textbf{19.825} & \textbf{0.797}/0.782 & \textbf{21.299}/21.218 & \textbf{0.842}/0.839 \\

      $\epsilon = 8$           & 22.108/\textbf{22.501} & \textbf{0.807}/0.803 & \textbf{20.999}/20.879 & 0.804/\textbf{0.805}
                  & 19.087/\textbf{19.207} & \textbf{0.754}/0.740 & \textbf{20.487}/20.427 & \textbf{0.813}/0.811 \\

  \hline
  \end{tabular}
\end{table*}
%-------------------------------------------------------------------------
%------------------------------------------------------------------------

%-------------------------------------------------------------------------
%------------------------------------------------------------------------
\begin{table*}
  \footnotesize
  \centering
  \caption{Quantitative defense results obtained by $L_{def}^{P}$ on D-Hazy. PSNR and SSIM are displayed in the form of A/B, where A is the result after defense training and B is the result before defense training, respectively. The bold value in A/B represents a better defense result.}
  \label{tab:defense_results_differ_attack_iterations}
  \begin{tabular}{c|c@{\hspace{0.75cm}}c|c@{\hspace{0.75cm}}c|c@{\hspace{0.75cm}}c|c@{\hspace{0.75cm}}c}
    \hline
      \multirow{2}{*}{Methods} & \multicolumn{2}{c|}{4KDehazing} & \multicolumn{2}{c|}{GCANet} & \multicolumn{2}{c|}{GridDehazeNet} & \multicolumn{2}{c}{MSBDN}\\
    \cline{2-9}
     & PSNR$\uparrow$ & SSIM$\uparrow$ & PSNR$\uparrow$ & SSIM$\uparrow$ & PSNR$\uparrow$ & SSIM$\uparrow$ & PSNR$\uparrow$ & SSIM$\uparrow$ \\
    
     \hline
  
      no attack           & 15.172/\textbf{23.304} & 0.747/\textbf{0.875} & 22.829/\textbf{24.209} & 0.838/\textbf{0.881} 
                           & \textbf{24.840}/24.649 & 0.890/\textbf{0.892} & 17.109/\textbf{23.093} & 0.779/\textbf{0.884}\\

      \hdashline

      $k = 1$       & \textbf{15.142}/14.534 & \textbf{0.693}/0.668 & \textbf{15.752}/12.088 & \textbf{0.733}/0.649 
                           & \textbf{15.687}/9.969 & \textbf{0.735}/0.599 & \textbf{16.921}/13.056 & \textbf{0.708}/0.639\\

      $k = 2$       & \textbf{15.131}/11.143 & \textbf{0.669}/0.596 & \textbf{15.457}/9.996  & \textbf{0.725}/0.547 
                           & \textbf{15.533}/9.346 & \textbf{0.729}/0.570 & \textbf{16.749}/9.703  & \textbf{0.683}/0.532\\

      $k = 3$       & \textbf{15.131}/10.057 & \textbf{0.658}/0.560 & \textbf{15.247}/9.307  & \textbf{0.718}/0.510 
                           & \textbf{15.453}/9.063 & \textbf{0.724}/0.548 & \textbf{16.702}/8.667  & \textbf{0.681}/0.493\\

      $k = 5$       & \textbf{15.135}/9.536  & \textbf{0.653}/0.534 & \textbf{15.057}/8.701  & \textbf{0.710}/0.477 
                           & \textbf{15.400}/8.647 & \textbf{0.721}/0.514 & \textbf{16.535}/7.997  & \textbf{0.693}/0.443\\

      $k = 10$      & \textbf{15.156}/8.918  & \textbf{0.658}/0.493 & \textbf{15.009}/8.148  & \textbf{0.708}/0.451 
                           & \textbf{15.372}/8.080 & \textbf{0.720}/0.459 & \textbf{16.289}/7.492  & \textbf{0.714}/0.421\\

      $k = 15$      & \textbf{15.164}/8.839  & \textbf{0.662}/0.482 & \textbf{15.001}/7.988  & \textbf{0.708}/0.442 
                           & \textbf{15.370}/7.830 & \textbf{0.720}/0.434 & \textbf{16.371}/7.413  & \textbf{0.719}/0.408\\

      $k = 20$      & \textbf{15.170}/8.462  & \textbf{0.665}/0.474 & \textbf{15.000}/7.837  & \textbf{0.708}/0.437 
                           & \textbf{15.365}/7.649 & \textbf{0.720}/0.417 & \textbf{16.340}/7.305  & \textbf{0.724}/0.399\\

      $k = 25$      & \textbf{15.174}/8.464  & \textbf{0.667}/0.460 & \textbf{15.003}/7.750  & \textbf{0.708}/0.432 
                           & \textbf{15.365}/7.538 & \textbf{0.720}/0.407 & \textbf{16.314}/7.239  & \textbf{0.725}/0.389\\

      $k = 30$      & \textbf{15.174}/8.397  & \textbf{0.668}/0.467 & \textbf{15.000}/7.725  & \textbf{0.708}/0.431 
                           & \textbf{15.363}/7.446 & \textbf{0.720}/0.398 & \textbf{16.262}/7.285  & \textbf{0.727}/0.396\\
      
  \hline
  \end{tabular}
\end{table*}
%-------------------------------------------------------------------------
%------------------------------------------------------------------------

%-------------------------------------------------------------------------
%------------------------------------------------------------------------
\begin{table*}
  \scriptsize
  \centering
  \caption{Quantitative defense results obtained by $L_{def}^{P}$ and $L_{def}^{G}$ on test datasets. PSNR and SSIM are displayed in the form of A/B, where A is the result after defense training and B is the result before defense training. The bold value in A/B (when $\epsilon > 0$) represents a better defense result.}
  \label{tab:defense_results_adv_training}
  \begin{tabular}{c|c@{\hspace{0.9cm}}c|c@{\hspace{0.9cm}}c|c@{\hspace{0.9cm}}c|c@{\hspace{0.9cm}}c}
    \hline
      \multirow{3}{*}{Methods} & \multicolumn{4}{c|}{TP} & \multicolumn{4}{c}{TG} \\
      \cline{2-9}
      & \multicolumn{2}{c|}{D-Hazy} & \multicolumn{2}{c|}{Foggy-City} & \multicolumn{2}{c|}{D-Hazy} & \multicolumn{2}{c}
      {Foggy-City} \\

    \cline{2-9}
     & PSNR$\uparrow$ & SSIM$\uparrow$ & PSNR$\uparrow$ & SSIM$\uparrow$ & PSNR$\uparrow$ & SSIM$\uparrow$ & PSNR$\uparrow$ & SSIM$\uparrow$ \\
    
     \hline   
      4KDehazing-0       & 15.250/\textbf{23.304} & 0.745/\textbf{0.875} & 24.533/\textbf{31.461} & 0.912/\textbf{0.974} 
      & 15.532/\textbf{23.304} & 0.753/\textbf{0.875} & 24.458/\textbf{31.461} & 0.909/\textbf{0.974}\\

      \hdashline

      4KDehazing-2       & \textbf{15.313}/10.302 & \textbf{0.739}/0.558 & \textbf{24.413}/15.626 & \textbf{0.906}/0.817 
      & \textbf{15.142}/9.363  & \textbf{0.732}/0.437 & \textbf{23.390}/15.273 & \textbf{0.893}/0.792\\

      4KDehazing-4       & \textbf{15.317}/9.000  & \textbf{0.705}/0.504 & \textbf{23.918}/13.538 & \textbf{0.884}/0.739 
      & \textbf{14.753}/7.574  & \textbf{0.704}/0.324 & \textbf{22.227}/13.142 & \textbf{0.868}/0.711\\

      4KDehazing-6       & \textbf{15.248}/8.475  & \textbf{0.648}/0.448 & \textbf{23.193}/12.768 & \textbf{0.850}/0.683 
      & \textbf{14.374}/6.983  & \textbf{0.672}/0.293 & \textbf{21.168}/12.567 & \textbf{0.837}/0.667\\

      4KDehazing-8       & \textbf{15.115}/8.061  & \textbf{0.576}/0.426 & \textbf{22.501}/12.355 & \textbf{0.803}/0.634 
      & \textbf{14.005}/6.713  & \textbf{0.636}/0.278 & \textbf{20.256}/12.250 & \textbf{0.800}/0.625\\

    \hline
    \hline
     
      GCANet-0           & 22.974/\textbf{24.209} & 0.843/\textbf{0.881} & 30.063/\textbf{31.541} & 0.957/\textbf{0.968} 
      & 22.655/\textbf{24.209} & 0.836/\textbf{0.881} & 30.335/\textbf{31.541} & 0.958/\textbf{0.968}\\

      \hdashline

      GCANet-2           & \textbf{15.463}/9.665  & \textbf{0.736}/0.557 & \textbf{23.486}/14.901 & \textbf{0.895}/0.787 
      & \textbf{15.358}/9.505  & \textbf{0.727}/0.545 & \textbf{23.461}/14.921 & \textbf{0.894}/0.785\\

      GCANet-4           & \textbf{15.074}/8.20   & \textbf{0.712}/0.458 & \textbf{22.581}/12.006 & \textbf{0.872}/0.652 
      & \textbf{14.922}/8.052  & \textbf{0.701}/0.443 & \textbf{22.480}/12.107 & \textbf{0.871}/0.653\\

      GCANet-6           & \textbf{14.645}/7.395  & \textbf{0.683}/0.392 & \textbf{21.701}/10.931 & \textbf{0.842}/0.568 
      & \textbf{14.475}/7.283  & \textbf{0.671}/0.381 & \textbf{21.562}/11.050 & \textbf{0.841}/0.573\\

      GCANet-8           & \textbf{14.253}/6.932  & \textbf{0.650}/0.354 & \textbf{20.879}/10.433 & \textbf{0.805}/0.516 
      & \textbf{14.070}/6.822  & \textbf{0.639}/0.346 & \textbf{20.722}/10.538 & \textbf{0.806}/0.522\\
      
    \hline
    \hline

      GridDehazeNet-0    & 24.261/\textbf{24.649} & 0.869/\textbf{0.892} & 29.970/\textbf{30.508} & 0.957/\textbf{0.961} 
      & \textbf{24.834}/24.649 & 0.885/\textbf{0.892} & 29.753/\textbf{30.508} & 0.956/\textbf{0.961}\\

      \hdashline

      GridDehazeNet-2    & \textbf{14.795}/9.332  & \textbf{0.722}/0.585 & \textbf{21.059}/15.279 & \textbf{0.850}/0.787 
      & \textbf{15.626}/9.316  & \textbf{0.732}/0.585 & \textbf{20.773}/15.082 & \textbf{0.855}/0.781\\

      GridDehazeNet-4    & \textbf{14.705}/8.067  & \textbf{0.708}/0.459 & \textbf{20.406}/14.300 & \textbf{0.819}/0.718 
      & \textbf{15.402}/8.028  & \textbf{0.713}/0.457 & \textbf{20.144}/13.924 & \textbf{0.829}/0.704\\

      GridDehazeNet-6    & \textbf{15.085}/7.268  & \textbf{0.690}/0.389 & \textbf{19.825}/13.521 & \textbf{0.782}/0.654 
      & \textbf{15.114}/7.226  & \textbf{0.687}/0.388 & \textbf{19.579}/13.220 & \textbf{0.795}/0.641\\

      GridDehazeNet-8    & \textbf{14.824}/6.753  & \textbf{0.656}/0.356 & \textbf{19.207}/12.932 & \textbf{0.740}/0.597 
      & \textbf{14.690}/6.714  & \textbf{0.653}/0.354 & \textbf{19.011}/12.700 & \textbf{0.753}/0.586\\

    \hline
    \hline

      MSBDN-0            & 22.771/\textbf{23.093} & 0.865/\textbf{0.884} & 29.370/\textbf{30.382} & 0.964/\textbf{0.969} 
      & 22.399/\textbf{23.093} & 0.864/\textbf{0.884} & 29.671/\textbf{30.382} & 0.964/\textbf{0.969}\\

      \hdashline

      MSBDN-2            & \textbf{16.068}/9.510  & \textbf{0.746}/0.561 & \textbf{23.380}/15.824 & \textbf{0.887}/0.788 
      & \textbf{15.723}/8.827  & \textbf{0.739}/0.485 & \textbf{23.421}/15.411 & \textbf{0.888}/0.770\\

      MSBDN-4            & \textbf{15.428}/7.422  & \textbf{0.715}/0.401 & \textbf{22.211}/12.874 & \textbf{0.865}/0.643 
      & \textbf{15.063}/6.858  & \textbf{0.704}/0.330 & \textbf{22.139}/12.277 & \textbf{0.866}/0.586\\

      MSBDN-6            & \textbf{14.836}/6.594  & \textbf{0.676}/0.333 & \textbf{21.218}/11.814 & \textbf{0.839}/0.555 
      & \textbf{14.469}/6.156  & \textbf{0.665}/0.261 & \textbf{21.101}/11.216 & \textbf{0.839}/0.492\\

      MSBDN-8            & \textbf{14.311}/6.215  & \textbf{0.634}/0.296 & \textbf{20.427}/11.148 & \textbf{0.811}/0.491 
      & \textbf{13.950}/5.738  & \textbf{0.624}/0.226 & \textbf{20.269}/10.713 & \textbf{0.812}/0.442\\
  \hline
  \end{tabular}
\end{table*}
%-------------------------------------------------------------------------
%------------------------------------------------------------------------

\subsection{Ablation Study and Discussions}
\label{subsec:ablation_study_and_discussions}

\subsubsection{Noise Attack and Gradient-based Attack}

The perturbation generated by the attacker can be seen as a specific noise information that is added to the hazy image. Therefore, we need to compare the attack results between the perturbation and the general noise to verify the effectiveness of the adversarial attack. The noise attack is marked as $A_{N}$, where $N$ denotes the noise sampling from noise distribution. It is worth noting that $A_{N}$ directly generates the $\delta$ by noise distribution, rather than using the gradient descent algorithm. The $A_{N}$ is mainly used to prove that the learned perturbation has stronger attack effects than general noise. Since the perturbation is initialized with a uniform distribution, the noise is also sampled from a uniform distribution. As our experimental results show, $L_{I}^{MSE}$ is the attack loss function with the overall worst attack performance in the quantitative evaluation. We compare noise attack $A_{N}$ and haze-preserved attack $L_{I}^{MSE}$ under the same experimental setting. The noise level of $A_{N}$ is set to $[-8, 8]$, which is consistent with the $\epsilon$ of $L_{I}^{MSE}$.

Table \ref{tab:attack_results_epsilon8_noise_and_haze-preserved} shows the attack performance of $A_{N}$ and $L_{I}^{MSE}$ against different dehazing methods. The results on all datasets demonstrate that the performance of noise attack is significantly lower than that of $L_{I}^{MSE}$. The visual results in Figure \ref{fig:noise_cityscapes} show the original dehazed outputs ($\epsilon = 0$) and the predicted attacked outputs ($\epsilon = 8$). Although $\epsilon$ is set to 8, the texture and structure of attacked outputs are still consistent with the original dehazed outputs. It can be concluded that all the attack algorithms proposed in this paper are significantly better than noise in attacking dehazing networks in terms of both quantitative and visual evaluation. The reason is that the $\delta$ is generated by adjusting the distance between $J_{p}^{\delta}$ and $X$. Moreover, the generation process of the $\delta$ utilizes the specific dehazing model. However, these two factors are not available in regular noise. Therefore, the attack effect of $\delta$ is stronger than regular noise.

\subsubsection{Attack without Ground-truth}

Table \ref{tab:attack_results_epsilon8_P_and_G} shows the attack results of $L_{G}^{MSE}$, in which the ground truth haze-free label is used in the attack process. The $\epsilon$ is set to $8$. Quantitative evaluation results show that the overall attack performance of $L_{G}^{MSE}$ is slightly better than that of $L_{P}^{MSE}$. It is reasonable that the attack effect of $L_{G}^{MSE}$ is slightly better than that of $L_{P}^{MSE}$ since the reference used in the calculation of quantitative metrics is the ground-truth haze-free image. The visual dehazed results $J_{p}^{\delta}$ obtained by attack loss $L_{G}^{MSE}$ are shown in Figure \ref{fig:cityscapes_gt}. The attack visual performances of Figure \ref{fig:cityscapes_gt} are quite close to the attack visual performance obtained by attack loss $L_{P}^{MSE}$ that is shown in Figure \ref{fig:pred_mse_cityscapes}. The quantitative and visual results demonstrate that taking predicted dehazed images ($J_{p}$) as pseudo labels is a promising way to attack the dehazing networks.

\subsubsection{Segmentation Results after Defense Training}

\begin{figure}
  \small
  \centering
  \includegraphics[width=8cm,height=4cm]{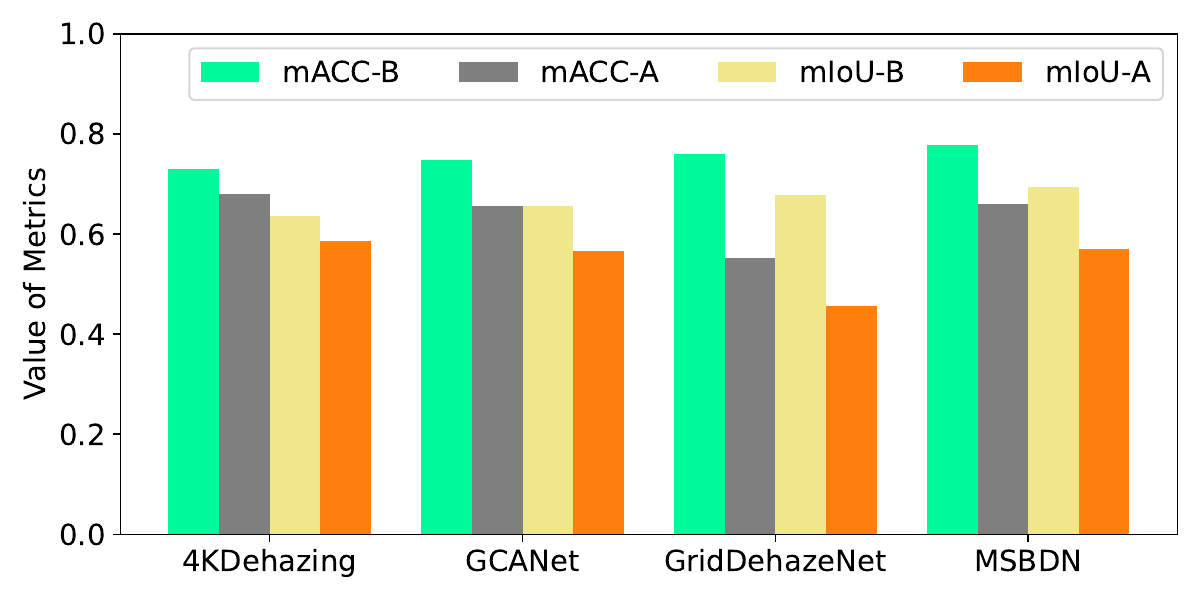}
  \caption{Attack results ($L_{P}^{MSE}$) on semantic segmentation dataset Foggy-City after defense training, where mACC and mIoU are two metrics for semantic segmentation. ``-B'' means before attack, and ``-A'' is after attack.}
  \label{fig:attack_quantitative_seg_cityscapes_after_defense}
\end{figure}

Figure \ref{fig:attack_quantitative_seg_cityscapes_before_defense} shows the segmentation results under attack $L_{P}^{MSE}$ when $\epsilon$ is set to $4$. Figure \ref{fig:attack_quantitative_seg_cityscapes_after_defense} gives the quantitative segmentation results after adversarial defense training and shows two interesting phenomenons. First, the segmentation results when there is no attack ($\epsilon = 0$) in Figure \ref{fig:attack_quantitative_seg_cityscapes_after_defense} are lower than that in Figure \ref{fig:attack_quantitative_seg_cityscapes_before_defense}. This is consistent with the dehazing results ($\epsilon = 0$) before and after defense training. Comparing Figure \ref{fig:attack_quantitative_seg_cityscapes_before_defense} with Figure \ref{fig:attack_quantitative_seg_cityscapes_after_defense}, we can find that the robustness of segmentation has been improved to a certain extent after defense training.

%-------------------------------------------------------------------------
%------------------------------------------------------------------------
%-------------------------------------------------------------------------
%------------------------------------------------------------------------

%-------------------------------------------------------------------------
%------------------------------------------------------------------------

\section{Conclusion}
In this paper, we try to explore and define a new research problem, which is called AADN. A first order attack algorithm is found to be powerful to attack the dehazing networks, which will lead to the performance reduction of the dehazing task. Meanwhile, several attack forms are proposed and verified for different attack purposes. Finally, we explore the effectiveness of adversarial defense training to protect vulnerable dehazing networks. We have investigated the fundamental problems of attacking dehazing networks. Since we must control the size of this paper, more attack and defense methods on the issue of the security of dehazing network will be studied in the future.

%-------------------------------------------------------------------------
%------------------------------------------------------------------------

\section*{Acknowledgments}
We thank the Big Data Computing Center of Southeast University for providing the facility support on the numerical calculations in this paper. This work was supported in part by National Key R\&D Program of China under Grant 2022YFB3104301; the grant of the National Science Foundation of China under Grant 62172090; the Fundamental Research Funds for the Central Universities; CAAI-Huawei MindSpore Open Fund. All correspondence should be directed to Xiaofeng Cong.

% \bibliographystyle{IEEEtran}
% \bibliography{egbib_full_name}

% \clearpage
% \clearpage

\end{document}